\documentclass[10pt,twocolumn,letterpaper]{article}

\usepackage{iccv}              

\newcommand{\PAR}[1]{\noindent{\bf #1}}
\newcommand*{\ourdataset}{EvLowLight\@\xspace}
\newcommand*{\ourmethod}{\textsc{RetinEV}\@\xspace}

\newcommand{\name}{0}

\usepackage{amssymb}  
\usepackage{pifont}   
\newcommand{\cmark}{\ding{52}}%
\newcommand{\xmark}{\ding{56}}%
\usepackage{colortbl}
\usepackage{diagbox}
\usepackage{adjustbox}
\usepackage{multirow}
\usepackage{subcaption}
\usepackage{mwe} 
\usepackage{dsfont}

\definecolor{tableHeadGray}{gray}{.9}
\definecolor{tableSubHeadGray}{gray}{0.95}
\newcommand{\pub}[1]{{\color{gray}{\tiny{[{#1}]\!}}}}
\usepackage{arydshln} 

\definecolor{iccvblue}{rgb}{0.21,0.49,0.74}
\usepackage[pagebackref,breaklinks,colorlinks,allcolors=iccvblue]{hyperref}

\def\paperID{1946} 
\def\confName{ICCV}
\def\confYear{2025}

\title{Low-Light Image Enhancement using Event-Based Illumination Estimation}

\author{%
Lei Sun\textsuperscript{1}\space\space\space\space
Yuhan Bao\textsuperscript{2}\space\space\space\space
Jiajun Zhai\textsuperscript{2}\space\space\space\space
Jingyun Liang\textsuperscript{3}\space\space\space\space
Yulun Zhang\textsuperscript{4}\space\space\space\space
Kaiwei Wang\textsuperscript{2}\\
Danda Pani Paudel\textsuperscript{1}\space\space\space\space
Luc Van Gool\textsuperscript{1}\vspace{6px}\\
\textsuperscript{1}INSAIT, Sofia University\space\space\space\space
\textsuperscript{2}Zhejiang University\space\space\space\space
\textsuperscript{3}Alibaba Group\space\space\space\space
\textsuperscript{4}Shanghai Jiao Tong University
}

\begin{document}


\maketitle

\begin{abstract}
    Low-light image enhancement (LLIE) aims to improve the visibility of images captured in poorly lit environments. 
    Prevalent event-based solutions primarily utilize events triggered by motion, \ie, ``motion events'' to strengthen only the edge texture, while leaving the high dynamic range and excellent low-light responsiveness of event cameras largely unexplored.
    This paper instead opens a new avenue from the perspective of estimating the illumination using ``temporal-mapping'' events, \ie, by converting the timestamps of events triggered by a transmittance modulation into brightness values.
    The resulting fine-grained illumination cues facilitate a more effective decomposition and enhancement of the reflectance component in low-light images through the proposed Illumination-aided Reflectance Enhancement module.
    Furthermore, the degradation model of temporal-mapping events under low-light condition is investigated for realistic training data synthesizing. 
    To address the lack of datasets under this regime, we construct a beam-splitter setup and collect \ourdataset dataset that includes images, temporal-mapping events, and motion events.
    Extensive experiments across 5 synthetic datasets and our real-world \ourdataset dataset substantiate that the devised pipeline, dubbed \ourmethod, excels in producing well-illuminated, high dynamic range images, outperforming previous state-of-the-art event-based methods by up to 6.62 dB, while maintaining an efficient inference speed of 35.6 frame-per-second on a $640\times480$ image. 
    

    \end{abstract}


    \section{Introduction}
    \label{sec:intro}
    Images captured under insufficient lighting, suffer from reduced visibility, resulting in an unsatisfactory visual experience and a decline in performance of computer vision systems optimized for well-lit images~\cite{li2021low}. Addressing this issue, low-light image enhancement (LLIE) has become a pivotal yet challenging task~\cite{wang2019underexposed,guo2016lime,cai2023retinexformer,hou2024global,wei2018deep}. The primary objective is to enhance visibility, Signal-to-Noise Ratio (SNR), and contrast for images taken under suboptimal illumination, while mitigating various distortions such as noise, artifacts, and color inaccuracies.
    Beyond image sensors, researchers have explored leveraging additional modalities to enhance outcomes further.
    {In this paper, we aim to utilize event cameras for the same.}

    \begin{figure}[t]
        \centering
        \includegraphics[width=0.98\linewidth]{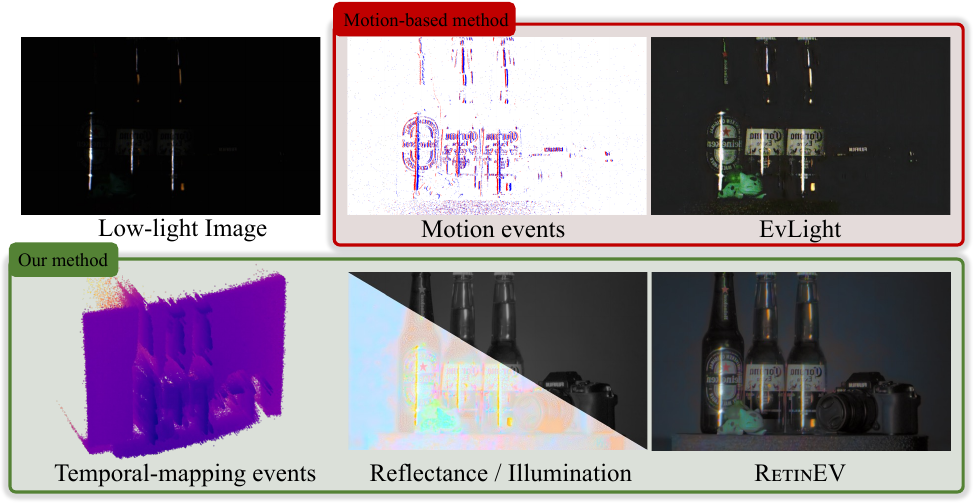}
        \caption{In contrast to existing event-based LLIE methods that rely on motion events~\cite{liang2024towards}, our \ourmethod leverages temporal mapping events for illumination estimation and reflectance decomposition, leading to better visibility.}
            
        \label{fig:teaser}
    \end{figure}

    Event cameras, also known as Dynamic Vision Sensors (DVSs), are neuromorphic sensors that respond to light changes~\cite{brandli2014240,gallego2020event,patrick2008128x,mead2023neuromorphic}, capturing ``\textit{event}'' information, at high temporal resolution (in the order of $\mu$s). Different from conventional frame-based cameras, the asynchronous architecture of event cameras endows them with high dynamic range (HDR) and exceptional low-light responsiveness. Events are triggered if change occurs in the scene, even under insufficient illumination, providing clues for LLIE. Prevalent event-based methods~\cite{jiang2023event,liang2024towards} focus on fusing low-light image with events, in which the information derived from \textit{motion events} plays a crucial role in enhancing low-light images. We will refer to these approaches as \textit{motion-based} methods.

    However, motion-based methods suffer from several limitations:
    \textbf{First}, In low-light environment, where longer exposure times are generally required, motion-based methods still rely on camera motion to generate events, \textit{resulting in degradations combining low-light effects and motion blur}.
    \textbf{Second}, events in motion-based methods only deliver edge information, which is \textit{insufficient for LLIE tasks that require fine-grained illumination estimation at both edges and within regions}. \textbf{Third}, these methods are also prone to generating \textit{artifacts}, as shown in Fig.~\ref{fig:teaser} (a).
    Hence, the full potential of event cameras in practical low-light image enhancement remains to be explored.

    Recent advance in event-to-image conversion task~\cite{bao2024temporal} produces impressive high-quality images from only events, which can be ascribed to triggering events with transmittance modulation instead of motion in the scene. This motivates us to rethink the current \textit{de facto} paradigm in event-based LLIE: Can we utilize events triggered by transmittance modulation, \ie, the temporal-mapping events for LLIE? Intrinsically, illumination\footnote{The CMOS sensor of a camera actually measures irradiance rather than illuminance, and in this paper, any mention of ``illuminance'' specifically refers to irradiance.} information encoded in temporal-mapping events is expected to offer higher density and illumination precision than motion events.
    
    With this insight, we propose a \textit{fast} technique, dubbed \ourmethod, that does not need scene motion to enhance low-light RGB images using event cameras. We capture temporal-mapping events of the scene, by changing the transmittance of the optical system. 
    These `temporal mapping events' refer to the brightness changes that are induced when the transmittance changes (opening a shutter in our case), and are harnessed to enhance the low-light RGB image by exploiting the Retinex theory~\cite{land1971retinex}. 
    This theory posits that an observed image results from the product of reflectance and illumination, where reflectance represents an intrinsic characteristic that remains constant under varying illumination conditions. 
    Illumination component is estimated from temporal mapping events, while also aids in the decomposition of the reflectance component, as shown in 
    Fig.~\ref{fig:teaser} (b). A coefficient is designed
    for arbitrary illumination manipulation in practical use, and the characteristics of events under low light are also considered in the degradation model for the model training on synthetic data.
    Unlike previous Retinex-based methods that focus solely on enhancing the illumination component~\cite{wu2022uretinex5,cai2023retinexformer,wei2018deep,yi2023diff}, our approach refines the reflectance component leveraging high-quality illumination components as prior information through a cross-modal attention mechanism.

    To instantiate our \ourmethod pipeline, we build a beam-splitter system consisting of a DVS and an RGB image sensor to capture the aligned events and low-light images. During data capture, the opening process of the mechanical shutter, which exposes the image sensor, also produces temporal-mapping events. This allows for the synchronization of image exposure and the capture of temporal-mapping events, completed as fast as 2 $m$s. 
    With this setup, we construct a real-world dataset named \ourdataset, comprising data captured under 60 extremely low illumination and high contrast scenes.
    The proposed method shows consistently superiority over state-of-the-art LLIE methods on five synthetic datasets and our real-world \ourdataset dataset, across objective reference and no-reference metrics, as well as subjective evaluations from user studies. The simple yet effective architecture also endows \ourmethod with a real-time inference speed of 35.6 frame-per-second on a 640$\times$480 image. The impressive results shed light on the promises of event cameras in LLIE.
    

    In a nutshell, we make the following main contributions:
    
    \begin{itemize}
        \item Based on Retinex theory, we propose a new and more effective paradigm named \ourmethod for practical event-based low-light image enhancement, where low-light event characteristics are considered and resulted brightness can be controlled.
        \item An Illumination-aid Reflectance Enhancement module is designed, where the illumination component is introduced to enhance reflectance with cross-modal attention.
        \item We construct a beam-splitter system and present \ourdataset, a real-world dataset consisting of aligned events and images under challenging illumination conditions, which provides an evaluation setting for LLIE methods.
        \item Quantitative and qualitative experiments show that \ourmethod outperforms image-only methods as well as earlier state-of-the-art event-based methods with real-time inference speed.
    \end{itemize}

    \section{Related Work}
    \label{sec:related_work}
    \PAR{Frame-Only Low-Light Image Enhancement.}
    \label{subsec:related_work_LIIE}
    Conventional LLIE methods fall into two main categories: histogram equalization~\cite{arici2009histogram1,nakai2013color2} and Retinex-based approaches~\cite{land1971retinex,jobson1997multiscale,guo2016lime}. Histogram equalization enhances contrast by adjusting intensity distributions~\cite{pizer1987adaptive}, while Retinex-based methods decompose images into illumination and reflectance, enabling illumination correction at different scales~\cite{land1971retinex,jobson1997multiscale}. However, these methods struggle with real-world variations and noise.
    Deep learning methods have significantly advanced LLIE. RetinexNet~\cite{wei2018deep} and its variants~\cite{wu2022uretinex5,yang2021sparse} use CNNs to decompose images and enhance illumination maps. Other approaches leverage illumination estimation~\cite{wang2019underexposed}, semantic priors~\cite{wu2023learning}, and signal-to-noise ratio modeling~\cite{xu2022snr4}, \etc. Despite their effectiveness, CNN-based methods~\cite{wang2019underexposed,wu2023learning,xu2022snr4,xu2023low7,zhang2019kindling13} often suffer from limited generalization under varying lighting conditions.
    Recent transformer-based models~\cite{liu2021swin,zamir2022restormer,liang2021swinir} have demonstrated improved performance on benchmark datasets. Meanwhile, generative models such as EnlightenGAN~\cite{jiang2021enlightengan}, Diff-Retinex~\cite{yi2023diff}, GSAD~\cite{hou2024global}, and DiffLL~\cite{jiang2023low} leverage GANs and diffusion models to enhance perceptual quality. However, these methods often introduce increased computational complexity.

    \begin{figure*}[t]
        \centering
        \includegraphics[width=0.96\linewidth]{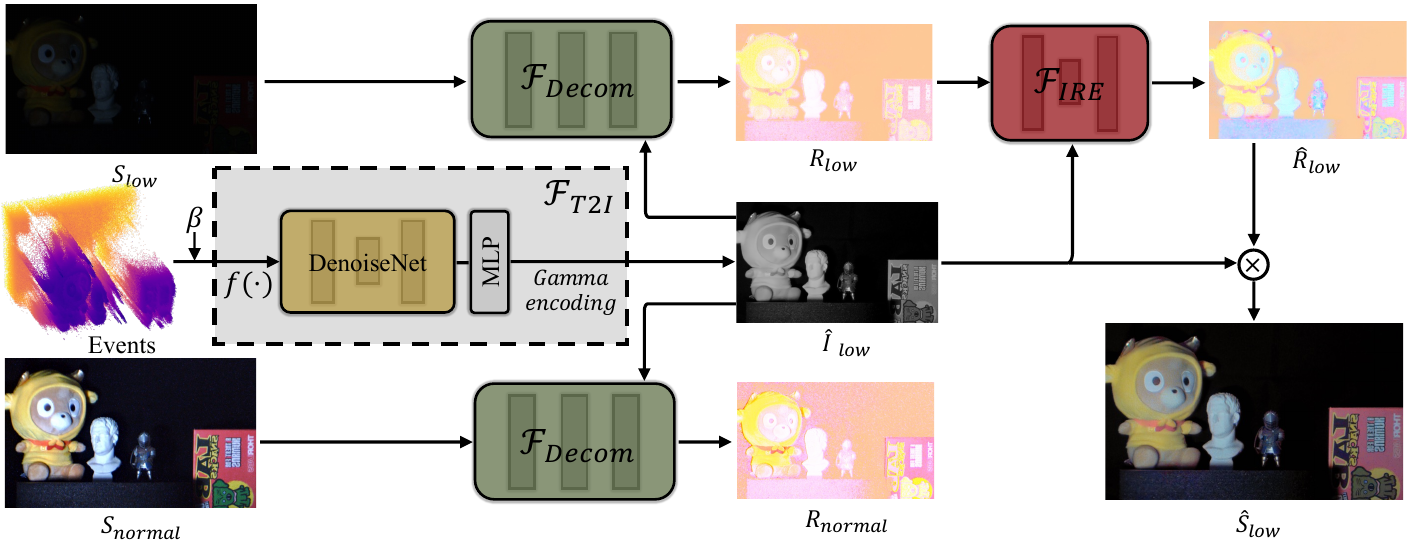}
        \vspace{-9pt}
        \caption{\textbf{The architecture of our \ourmethod.} $\mathcal{F}_{Decom}$ for low-light images and normal-light images share the same weights. $\beta$ is added for brightness manipulation. ``IRE'': Illumination-aid Reflectance Enhancement, ``$S_{low}$'': low-light image, ``$S_{normal}$'': normal-light image, ``$I$'': illumination component, ``$R$'': Reflectance component, ``$\hat{S}_{low}$'': the predicted enlightened image.}
        \vspace{-15pt}
        \label{fig:framework}
    \end{figure*}

    \PAR{Event-Based Image Enhancement}
    \label{subsec:event_based_image_enhancement}
     employs a hybrid sensor model, where event data aids image quality improvement, as in our method. This approach has been explored in various tasks, including image deblurring~\cite{edi_pan,vitoria2022event,kim2021event,zhang2022unifying,shang2021bringing,haoyu2020learning,lin2020learning_event_video_deblur,sun2022event,zhou2023deblurring}, HDR imaging~\cite{Messikommer_2022_CVPR}, video frame interpolation~\cite{tulyakov2021time,sun2023event,tulyakov2022time}, and real-time photometric stereo~\cite{yu2024eventps}, \etc. These systems share a common focus on motion-related image enhancement.
    The HDR and low-light responsiveness of the event camera has inspired their use in low-light image and video enhancement. Jiang~\etal~\cite{jiang2023event} and Liang~\etal~\cite{liang2024towards} utilize motion-free low-light images and events near exposure times for enhancement. However, prolonged exposure in low-light conditions can blur motion-triggered events, and events primarily capture high-contrast edges without grayscale information, limiting their utility.
    Recent works~\cite{liu2023low,liang2023coherent} model video coherence using events, facilitating frame alignment and fusion. However, these methods primarily build upon existing image enhancement techniques by incorporating multi-frame alignment. As a result, the potential of event cameras in low-light image enhancement remains largely unexplored.

    \vspace{-3pt}
    \section{\ourmethod}
    \label{sec:method}

    We first establish the mathematical relationship between illumination and the timestamps of temporal-mapping events in \S\ref{subsec:preliminaries}. Leveraging the estimated illumination from events, we incorporate Retinex theory~\cite{land1971retinex} to construct the architecture of \ourmethod in \S\ref{subsec:architecture}. We then detail the T2I and IRE modules in \S\ref{subsec:T2I} and \S\ref{subsec:IRE}, respectively. Finally, loss functions are designed in \S\ref{subsec:details}.

    
    \subsection{Problem Formulation}
    \label{subsec:preliminaries}
    
    In event cameras, an event, typically represented as a tuple $e=(x,y,t,p)$, is triggered once the change in light intensity $\mathcal{I}$ exceeds a predefined threshold in the log domain:
    \begin{equation}
    \label{eq:event_trigger}
    p = 
    \begin{cases}
    +1, \text{if} \log \left( \frac{\mathcal{I}_t {(x, y)}} {\mathcal{I}_{ t - \Delta t} {(x, y)}} \right) > c, \\
    -1, \text{if} \log \left( \frac{\mathcal{I}_t {(x, y)}} {\mathcal{I}_{ t - \Delta t} {(x, y)}} \right) < -c,
    \end{cases}
    \end{equation}
    where $x$, $y$, $p$, $t$, $c$ represent the coordinates, polarity, timestamp, and the contrast threshold of the event, resp.
    
    
    Previous work~\cite{bao2024temporal} actively changes the transmittance of the optical system to modify the illuminance received by the sensor. It then uses the timestamp of the first positive event (FPE) generated at each pixel location to estimate the local illumination. In our case, we set the transmittance function to be a step function. The FPE of a pixel is triggered when the energy of the photoelectric conversion at the pixel reaches the threshold voltage, at which point the energy stored in the capacitor is reached, that is,
    \begin{equation}
        \label{eq:energy}
          \eta \cdot E \cdot A \cdot t_{fpe} = \frac{C \cdot U_{thd}^{2}} {2},
    \end{equation}
    where $\eta$, $E$, $A$, $t_{fpe}$, $C$, and $U_{thd}$ represents photoelectric conversion efficiency, illuminance, the photosensitive area of the pixel, timestamp of FPE, pixel capacitance, and pixel threshold voltage, resp. 
    
    From Eq.~(\ref{eq:energy}), the relationship between the illuminance $E$, the physical quantity to be measured, and the time $t_{fpe}$ can be expressed as a conversion function $f(\cdot)$:
    \begin{equation}
        \label{eq:E}
        E = \frac{C \cdot U_{thd}^{2}}{2 \eta \cdot A \cdot t_{fpe}} = \frac{k}{t_{fpe}} = f(t_{fpe}).
    \end{equation}
    
    
    Due to various complex factors, the illuminance $E$ in real-world scenarios \textit{cannot} be precisely determined using Eq.~(\ref{eq:E}). However, we can leverage the general relationship between $E$ and $t_{fpe}$ to construct our model, which will be introduced in the following section.

    \subsection{General Architecture}
    \label{subsec:architecture}
    According to the classic Retinex theory~\cite{land1971retinex}, an observed image $S$ can be decomposed into two components: reflectance component $R$ and illumination component $I$.
    \begin{equation}
        \label{eq:retinex}
        S = R \cdot I.
    \end{equation}
    $I$ is determined by the lighting conditions. 
    Prevalent image-only Retinex-based methods~\cite{cai2023retinexformer,wu2022uretinex5,wei2018deep} estimate the $I_{low}$ and $R_{low}$ from low-light images and enhance $I$. However, sensor response degradation in low light leads to suboptimal $I_{low}$ estimation, with \textit{defects further amplified} during illumination enhancement for $\hat{I}_{low}$.
    
    In \ourmethod (shown in Fig.~\ref{fig:framework}), temporal-mapping events with \textit{better low-light responsiveness} are harnessed for the estimation and enhancement of $I_{low}$. Note that different from well-defined physics, the Retinex theory is based on human perception. As a consequence, there is no precise conversion between illumination $I$ (subjective measure) and illuminance $E$ (objective physical quantity, derived from $t_{fpe}$ in Eq.~(\ref{eq:E})). Hence, we design a Time-to-Illumination (T2I) module for the conversion and the enhancement of illumination:
    \begin{equation}
        \label{eq:I_predict}
        \hat{I} = \mathcal{F}_{T2I}(t_{fpe}(x,y);\Theta_{1}), \quad \forall (x,y) \in \Omega.
    \end{equation}
    Here, $\Theta_1$, as well as $\Theta_2$ and $\Theta_3$ mentioned later in this section, represent the learnable parameters of the corresponding mappings. $\Omega$ represents the spatial domain of the image sensor. The T2I module consists of the conversion function $f(\cdot)$, a DenoiseNet, Multilayer Perceptron (MLP) layers for non-linear mapping, and Gamma encoding~\cite{debevec2004high} to ensure alignment with RGB images, as shown in Fig~\ref{fig:framework}.
    
    \begin{figure}[t]
        \centering
        \includegraphics[width=0.97\linewidth]{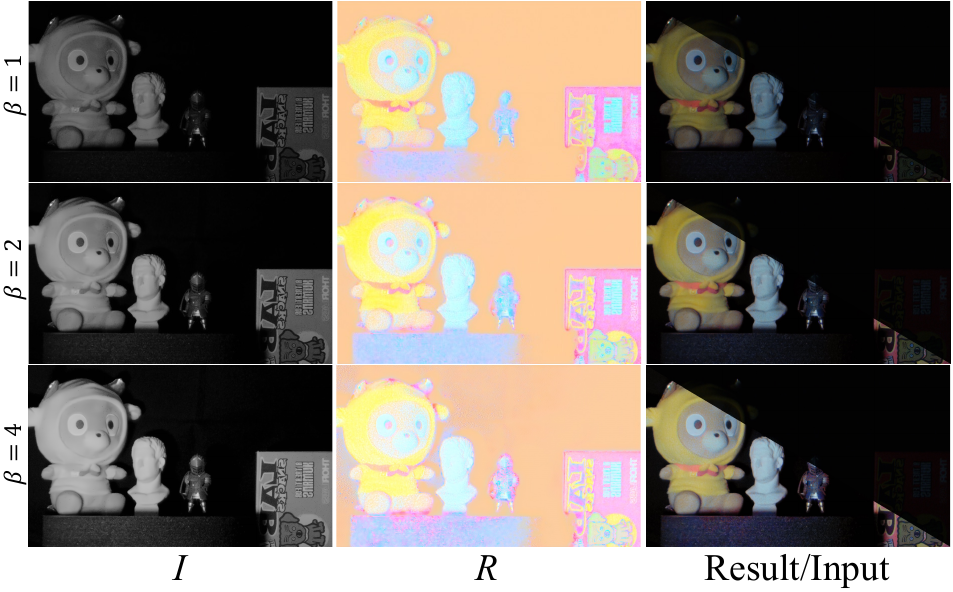}
        \vspace{-10pt}
        \caption{\textbf{Visualization of the effect of $\beta$ on the brightness of I and the result}.}
        \vspace{-20pt}
        \label{fig:beta}
    \end{figure}

    
    

    Regarding the reflectance component $R$, classical Retinex theory asserts that $R$ is an intrinsic property of the captured object and remains invariant under varying illumination conditions~\cite{land1971retinex,wei2018deep}. However, in practice, similar to the illumination component $I$ extracted from RGB camera under low-light conditions, the reflectance component $R$ also \textit{suffers severe degradation}. Prior studies have demonstrated that accurately estimating $I$ significantly facilitates the determination of $R$~\cite{cai2023retinexformer,wu2022uretinex5}. To this end, we introduce the well-estimated illumination $\hat{I}$ from \textit{events} into the decomposition process:
    \begin{equation}
        \label{eq:R_low}
        R_{low} = \mathcal{F}_{Decom}(S_{low}, \hat{I};\Theta_{2}),
    \end{equation}
    and subsequently enhance $R_{low}$ as follows:
    \begin{equation}
        \label{eq:R_hat_low}
        \hat{R}_{low} = \mathcal{F}_{IRE}(R_{low}, \hat{I}; \Theta_{3}).
    \end{equation}
    Here, the Illumination-aided Reflectance Enhancement (IRE) module $\mathcal{F}_{IRE}$ captures long-range dependencies between the enhanced reflectance $\hat{R}_{low}$ and the illumination estimate $\hat{I}$, as detailed further in Section~\ref{subsec:IRE}.



    Finally, the predicted $\hat{S}_{low}$ that approximates normal-light image $S_{normal}$ is derived by:
    \begin{equation}
        \label{eq:final}
        \hat{S}_{low} = \hat{I} \cdot \hat{R}_{low}.
    \end{equation}
    
    The general architecture of \ourmethod is shown in Fig.~\ref{fig:framework}. 
    

    \subsection{Time-to-Illumination Module}
    \label{subsec:T2I}
    To leverage the illumination prior, the Time-to-Illumination (T2I) module is designed to convert temporal events to illumination for normal-light image synthesis, as shown in Fig.~\ref{fig:framework}. Though event cameras exhibit superior low-light responsiveness, their performance still degrades under low-light conditions.
    
    \PAR{Low-Light Degradation Model (LLDM).}
    Different from the temporal-mapping events in EvTemMap~\cite{bao2024temporal}, which are captured under normal-light conditions, events in low-light environments exhibit distinct characteristics~\cite{whitepaper}, mainly including: (\textit{i}) longer latency and (\textit{ii}) increased dark current noise. We design and formulate LLDM to make synthetic data in the training process more closely approximate real-world data.
    In the training phase, (1) the GT images are first applied with degradation on the intensity of pixels, \ie, \textit{spatial domain}, including: \textit{blurriness} to synthesize the diffraction effects, \textit{downsampling} for the sensor limitation, and \textit{poisson-Gaussian} hybrid noise for the low-light dark current noise. 
    Then, the images are converted to the \textit{temporal domain}, introduced with \textit{timestamp latency}, \textit{dead pixel}, and \textit{random contrast threshold} degradation. Latency and the probability of dead pixels are directly proportional to the timestamp, \ie, a larger timestamp corresponds to higher latency and an increased probability of dead pixels.
    
    All the degradation is randomly sampled within a range, and randomly shuffled in each domain to cover real-world conditions. For more details kindly \cf §\ref{subsec:training_details} and \S\textcolor{iccvblue}{2} in supp.

    

    \PAR{Illumination Manipulation.} The preferred illumination intensities can vary significantly across individuals and applications. To make \ourmethod more practical, an illumination manipulation coefficient $\beta$ is introduced in the event-to-intensity conversion to adjust the illumination strengths:
    \begin{equation}
        \label{eq:beta}
        t_{norm} = \frac{t_{fpe}+\beta}{max(t_{fpe}) + \beta}.
    \end{equation}
    With $t_{norm}$, the value of $E$ is obtained using Eq.~(\ref{eq:E}). By choosing different $\beta$, we can adjust the intensity of $I$ as well as the result images, as shown in Fig.~\ref{fig:beta}. As the changing if $I$, $R$ should be invariant, which is also indicated in the visualized $R$.

    \subsection{Illumination-aid Reflectance Enhancement}
    \label{subsec:IRE}
    Image-only Retinex-based LLIE methods~\cite{wu2022uretinex5,wei2018deep,cai2023retinexformer} focus mainly on the enhancement of illumination. 
    For our method, with the high-quality illumination from temporal-mapping events, we improve the reflectance from the limited image sensor in IRE module.
    After layer normalization and 1$\times$1 convolution, queries ($\mathbf{Q}$), keys ($\mathbf{K}$) and values ($\mathbf{V}$) are derived. In our case, the $\mathbf{Q}$ is from $R$ component while $\mathbf{K}$ and $\mathbf{V}$ are from $I$ (denoted as $\mathbf{Q}_{{R}}$, $\mathbf{K}_{{I}}$, and $\mathbf{V}_{{I}}$, resp.). 
    Because $R$ and $I$ are the same size as the input image, this results in a substantial computational burden. Hence, different from vanilla attention mechanism~\cite{vaswani2017attention}, the $\mathbf{Q}_{{R}}$ is transposed before multi-head attention:
    \begin{equation} 
        \label{eq:EICA}
        \mathrm{Attention}(\mathbf{Q}_{{R}},\mathbf{K}_{{I}},\mathbf{V}_{{I}}) = \mathbf{V}_{{I}}\mathrm{softmax}\left(\frac{\mathbf{Q}^\intercal_{{R}}\mathbf{K}_{{I}}}{\sqrt{d_{k}}}\right),
    \end{equation}
    where $d_{k}$ denotes the dimensionality of the key (and query) vectors in each attention head. The attention map in Eq.~(\ref{eq:EICA}) is $c \times c$, reducing the spatial complexity from $\mathcal{O}(h^2w^2)$ to $\mathcal{O}(c^2)$, where $h$, $w$, and $c$ denotes height, width, and channel of the input feature maps, resp. Long-range dependency is better captured between two modalities while maintaining a low computational cost.

    \begin{figure}[t]
        \centering 
        \includegraphics[width=0.49\textwidth]{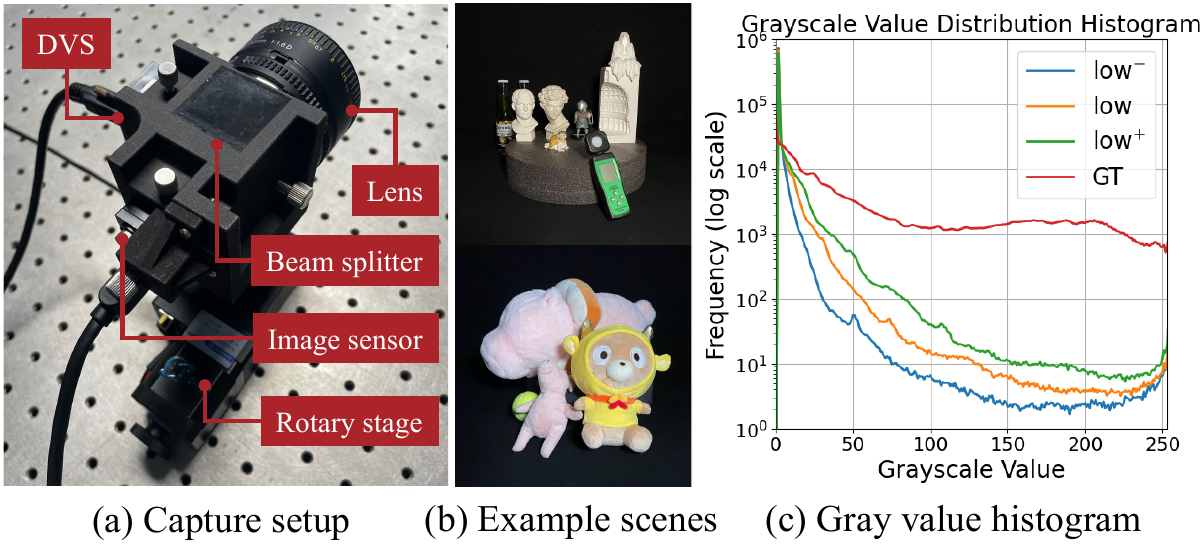}
        \caption{\textbf{Details} about \ourdataset dataset. The beam splitter is positioned between the lens and sensors.}
        \vspace{-8pt}
        \label{fig:dataset}
    \end{figure}

    \subsection{Loss Functions}
    \label{subsec:details}
    In our training stage, both low-light images and normal-light images are forwarded (shown in Fig.~\ref{fig:framework}). The normal-light image should be reconstructed by combining $\hat{I}$ from events and $R$ from both normal-light image $S_{normal}$ and low-light image $S_{low}$, resulting in a reconstruction loss:
    
    \begin{equation}
        \small
        \label{eq:loss_reconstruct}
        \mathcal{L}_{recon}= ||\hat{I} \cdot \hat{R}_{low} - S_{normal}||_{1} + ||\hat{I} \cdot {R}_{normal} - S_{normal}||_{1}.
    \end{equation}
    
    Following Retinex theory, the reflectance remains constant under varying illuminations, leading to the concept of an invariant reflectance loss:

    \begin{table}[t]
    \centering
    \small
    \setlength{\tabcolsep}{1pt}
    \renewcommand\arraystretch{1.1}
    \caption{\textbf{Comparison} with existing datasets.}
        \vspace{-10pt}
        \setlength{\tabcolsep}{4pt}  
        \renewcommand\arraystretch{1.1}
        \resizebox{0.48\textwidth}{!}{%
        \begin{tabular}{r||ccc}
        \bottomrule[0.15em]
        \rowcolor{tableHeadGray}
             & Motion events & Temporal-mapping events & Resolution \\ \hline \hline
         LOL v1~\pub{BMVC2018}~\cite{wei2018deep}  &  \xmark & \xmark & $600\times400$  \\
         SDSD~\pub{ICCV2021}~\cite{wang2021seeing}  &  \xmark & \xmark & $512\times960$  \\
         ELIE~\pub{TMM2023}~\cite{yu2023learning} &  \cmark & \xmark & $256\times256$  \\
         SDE~\pub{CVPR2024}~\cite{liang2024towards} &  \cmark & \xmark & $260\times346$  \\
        \textbf{\ourdataset (Ours)} &  \cmark & \cmark & $1216\times720$  \\
         \hline
        \end{tabular}}
        \label{tab:dataset_comparison}
    \vspace{-8pt}
    \end{table}

    \begin{equation}
        \label{eq:loss_requal}
        \mathcal{L}_{R} = ||R_{low} - R_{normal}||_1 +  ||\hat{R}_{low} - R_{normal}||_1.
    \end{equation}
    
    Finally, perceptual loss is added to guide the model toward generating outputs that are perceptually closer to the target, capturing finer textures and structural details:
    \begin{equation}
        \label{eq:loss_final}
        \mathcal{L}_{final}= \lambda_{1}\mathcal{L}_{recon} + \lambda_{2}\mathcal{L}_{R} + \lambda_{3}\mathcal{L}_{percep},
    \end{equation}
    where $\lambda_{1}$, $\lambda_{2}$, and $\lambda_{1}$ are hyper-parameters.

    \section{\ourdataset Dataset}
    \label{sec:dataset}
    
    \begin{table*}[th]
    \centering
    \small
    \setlength{\tabcolsep}{4pt}
    \renewcommand\arraystretch{1.1}
    \caption{\textbf{Details of \ourdataset dataset.} Image: image sensor, T.M. events: temporal-mapping events, M. events: motion events.}
    \vspace{-10pt}
    \resizebox{0.98\textwidth}{!}{%
        \begin{tabular}{cccccccccccc}
        \bottomrule[0.12em]
        \rowcolor{tableHeadGray}
          &  &  &  & \multicolumn{6}{c}{\textbf{Image exposure time in each group ($m$s)}} &  &  \\ \cline{5-10}
        \rowcolor{tableHeadGray}
         \multirow{-2}{*}{\textbf{Device}} & \multirow{-2}{*}{\textbf{Size}} & \multirow{-2}{*}{\textbf{Groups}} & \multirow{-2}{*}{\textbf{Illuminance}} & \textbf{$\mathrm{low^{-}}$} & \textbf{$\mathrm{low^{}}$} & \textbf{$\mathrm{low^{+}}$} & \textbf{$\mathrm{ldr^{-}}$} & \textbf{$\mathrm{ldr^{}}$} & \textbf{$\mathrm{ldr^{+}}$} & \multirow{-2}{*}{\textbf{T.M. events}} & \multirow{-2}{*}{\textbf{M. events}} \\ \hline
        DVS + Image & 1216$\times$720 & 60 & 2.5--6 lux & 5--8 & 10--15 & 15--20 & 120--180 & 200--400 & 400--700 & \cmark & \cmark \\ \hline
        \end{tabular}}
        \vspace{-6pt}
        \label{tab:dataset}
    \end{table*}

    \begin{table*}[th]
    \centering
    \small
    \setlength{\tabcolsep}{2pt}
    \renewcommand\arraystretch{1.1}
    \caption{\textbf{Quantitative comparisons} on LOL~\cite{wei2018deep,yang2020fidelity} (v1, v2-real, and v2-syn), and SDSD~\cite{wang2021seeing} (in and out) datasets. The best results are in \textbf{bold}.}
    \vspace{-10pt}
    \resizebox{0.98\textwidth}{!}{%
        \begin{tabular}{r||ccccccccccccccccc}
        \bottomrule[0.12em]
        \rowcolor{tableHeadGray}
        \textbf{Method}  & \textbf{~~Events} & &
        \textbf{~~~PSNR}  & \textbf{SSIM~~~} & & 
        \textbf{~~~PSNR} & \textbf{SSIM~~~}  & &
        \textbf{~~~PSNR} & \textbf{SSIM~~~}  & &
        \textbf{~~~PSNR} & \textbf{SSIM~~~}  & &
        \textbf{~~~PSNR} & \textbf{SSIM~~~}  &  \textbf{\#Params (M)} \\ \hline
         \hline
          &   &  & 
        \multicolumn{2}{c}{\textbf{LOL-v1}} &  & 
        \multicolumn{2}{c}{\textbf{LOL-v2-real}} & &
        \multicolumn{2}{c}{\textbf{LOL-v2-syn}} &  &
        \multicolumn{2}{c}{\textbf{SDSD-in}} & &
        \multicolumn{2}{c}{\textbf{SDSD-out}} & \\ \hline

        LIME~\pub{TIP2016}~\cite{guo2016lime}     & \xmark &  & 
        16.76 & 0.440 &  & 
        15.24  & 0.470 & & 
        16.88 & 0.776 & & 
        -  & - & & 
        -  & - & \textbf{0.77} \\

        RetinexNet~\pub{BMVC2018}~\cite{wei2018deep}     & \xmark &  & 
        16.77 & 0.560 &  & 
        20.06  & 0.816 & & 
        22.05 & 0.905 & & 
        23.25 & 0.863 & & 
        25.28 & 0.804 & {0.84} \\
        
        DeepUPE~\pub{CVPR2019}~\cite{wang2019underexposed}     & \xmark &  & 
        14.38 & 0.446 &  & 
        13.27  & 0.452 & & 
        15.08 & 0.632 & & 
        21.70 & 0.662 & & 
        21.94 & 0.698 & 1.02 \\
      
        KinD~\pub{MM2019}~\cite{zhang2019kindling13}     & \xmark &  & 
        20.86 & 0.790 &  & 
        14.74  & 0.641 & & 
        13.29 & 0.578 & & 
        21.95 & 0.672 & & 
        21.97 & 0.654 & 8.02 \\
    
        EnGAN~\pub{TIP2021}~\cite{jiang2021enlightengan}     & \xmark &  & 
        17.48 & 0.650 &  & 
        18.32  & 0.617 & & 
        16.57 & 0.734 & & 
        20.02 & 0.604 & & 
        20.10 & 0.616 & 114.35 \\

        Restormer~\pub{CVPR2022}~\cite{zamir2022restormer}     & \xmark &  & 
        22.43 & 0.823 &  & 
        19.94  & 0.827 & & 
        21.41 & 0.830 & & 
        25.67 & 0.827 & & 
        24.79 & 0.802 & 26.13 \\
    
        SNR-Net~\pub{CVPR2022}~\cite{xu2022snr4}     & \xmark &  & 
        24.61 & 0.842 &  & 
        21.48  & 0.849 & & 
        24.14 & 0.928 & & 
        29.44 & 0.894 & & 
        28.66 & 0.866 & 4.01 \\
        
        Retinexformer~\pub{ICCV2023}~\cite{cai2023retinexformer}     & \xmark &  & 
        25.16 & 0.845 &  & 
        22.80  & 0.840 & & 
        25.67 & 0.930 & & 
        29.77 & 0.896 & & 
        29.84 & 0.877 & 1.61 \\
        \arrayrulecolor{gray}\hdashline\arrayrulecolor{black}
        ELIE~\pub{TMM2023}~\cite{jiang2023event}     & \cmark &  & 
        - & - &  & 
        -  & - & & 
        - & - & & 
        27.46 & 0.879 & & 
        23.29 & 0.742 & 204.95 \\
        
        Liu~\etal~\pub{AAAI2023}~\cite{liu2023low}     & \cmark &  & 
        - & - &  & 
        -  & - & & 
        - & - & & 
        27.58 & 0.888 & & 
        23.51 & 0.726 & - \\
        
        EvLight~\pub{CVPR2024}~\cite{liang2024towards}     & \cmark &  & 
        - & - &  & 
        - & - & & 
        - & - & & 
        28.52 & 0.913 & & 
        26.67 & 0.836 & 22.73 \\
    
        \textbf{\ourmethod} (\textbf{Ours})     & \cmark &  & 
        \textbf{28.60} & \textbf{0.877} &  & 
        \textbf{30.32} & \textbf{0.929} & & 
        \textbf{32.06} & \textbf{0.951} & & 
        \textbf{33.65} & \textbf{0.960} & & 
        \textbf{33.29} & \textbf{0.958} & 3.44 \\
        
        \hline
        \end{tabular}}
        \label{tab:synthetic}
    \vspace{-5pt}
    \end{table*}

    \begin{figure*}[t]
        \centering
        \includegraphics[width=0.98\linewidth]{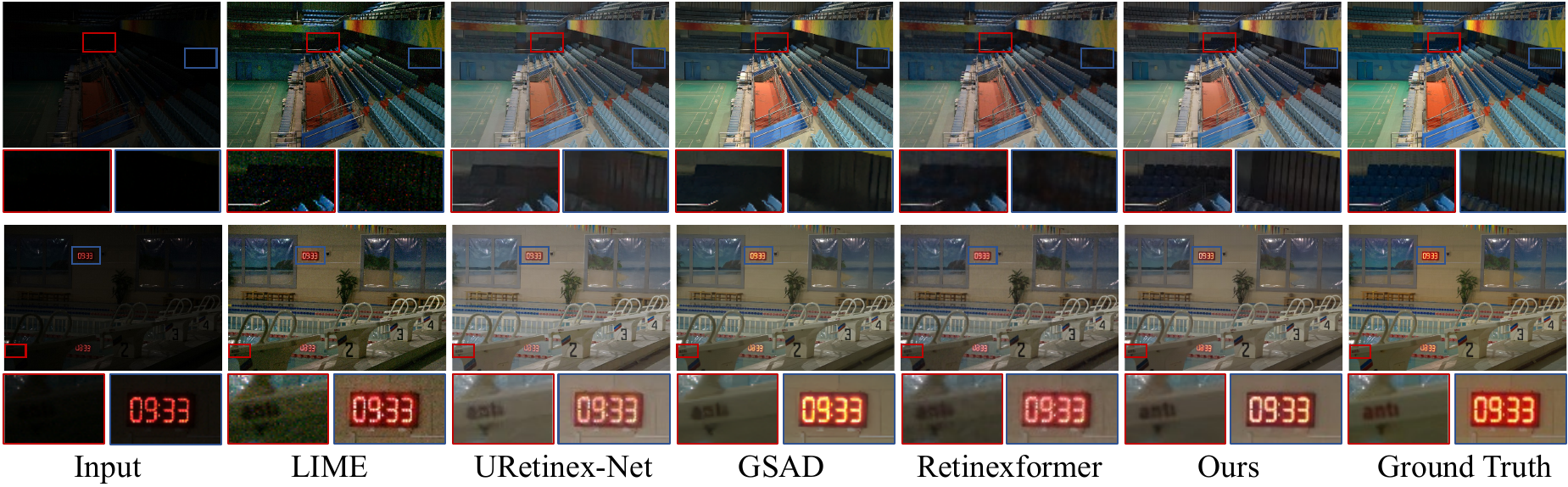}
        \vspace{-8pt}
        \caption{\textbf{Visual comparison on LOL-v1~\cite{wei2018deep} dataset}. Our method effectively enhances visibility and preserves fine-grained textures.}        \vspace{-10pt}
        \label{fig:qualitative_lolv1}
    \end{figure*}

    Our \ourmethod establishes a new paradigm for event-based LLIE. Existing datasets~\cite{liang2024towards,jiang2023event}, relying on motion events, are incompatible due to the absence of temporal-mapping events as shown in Table~\ref{tab:dataset_comparison}.
    
    Hence, in this work, we construct a share-lens beamsplitter system, as shown in Fig.~\ref{fig:dataset} (a). Different from previous dual-camera setup~\cite{tulyakov2021time} or beam-splitting-first style beamsplitter configurations~\cite{tulyakov2022time}, the 50R/50T beamsplitter in our setup is placed between the lens and sensors. This positioning provides both sensors with an imaging beam that preserves identical optical characteristics, enabling spatial alignment at any depth using the same homography matrix~\cite{hartley2003multiple}. 
    The Nikkor 50mm f/1.8D lens was specially modified with a leaf shutter to adjust the transmission of the optical system, enabling the generation of temporal-mapping events.
    Prophesee EVK4 ($1280\times720$) and MindVision MV-SUA134GC ($1280\times1024$) serves as DVS and image sensor, separately. The entire system is mounted within a rigid plastic casing, connected at the base to a rotary stage and secured to an optical platform. Grayscale value distribution is shown in Fig~\ref{fig:dataset} (c).
    
    Using the shared-lens beamsplitter setup, we collected a novel real-world event-based low-light image enhancement dataset, \ourdataset, detailed in Table~\ref{tab:dataset}. The dataset includes 60 scenes with illuminance levels ranging from 2.5 to 6 lux, with example scenes shown in Fig.~\ref{fig:dataset} (b).
    For each scene, we captured two sets of images: low-light images, and normal-light images. Each set consists of 3 images with short, regular and long exposure time, representing extreme-underexposure, normal, and slightly brighter low-light/normal-light conditions.
    Normal-light images with different exposure time are utilized for exposure fusion~\cite{mertens2007exposure} for producing reference HDR images.
    Thereafter, temporal-mapping events are recorded by opening the leaf shutter within as short as 2 $m$s. Since our image sensor does not support a mechanical shutter, temporal-mapping event acquisition could otherwise \textit{seamlessly} integrate with image exposure. Finally, we generated moving events by gently rotating the rotary stage, which served as input events for the event-based competitors~\cite{liang2024towards}. Note that \ourdataset is used only for evaluation because \ourmethod supports training on a synthetic dataset.
    
    
    \section{Experiments}
    \label{sec:experiments}

    \begin{figure*}[!t]
        \centering
        \setcounter{subfigure}{0} 
        \renewcommand{\name}{figures/qualitative_ourdataset/38low2/}
        \begin{subfigure}{0.195\textwidth}
            \includegraphics[width=\linewidth]{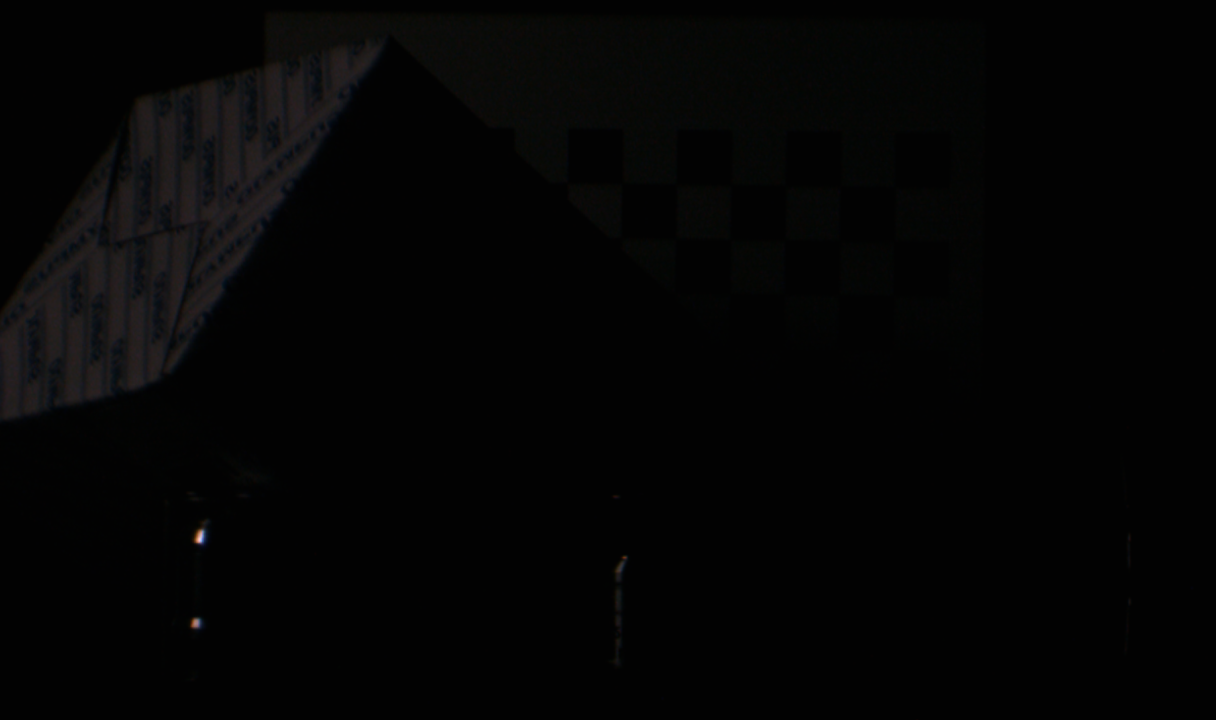}
            \caption*{Input}
        \end{subfigure}
        \begin{subfigure}{0.195\textwidth}
            \includegraphics[width=\linewidth]{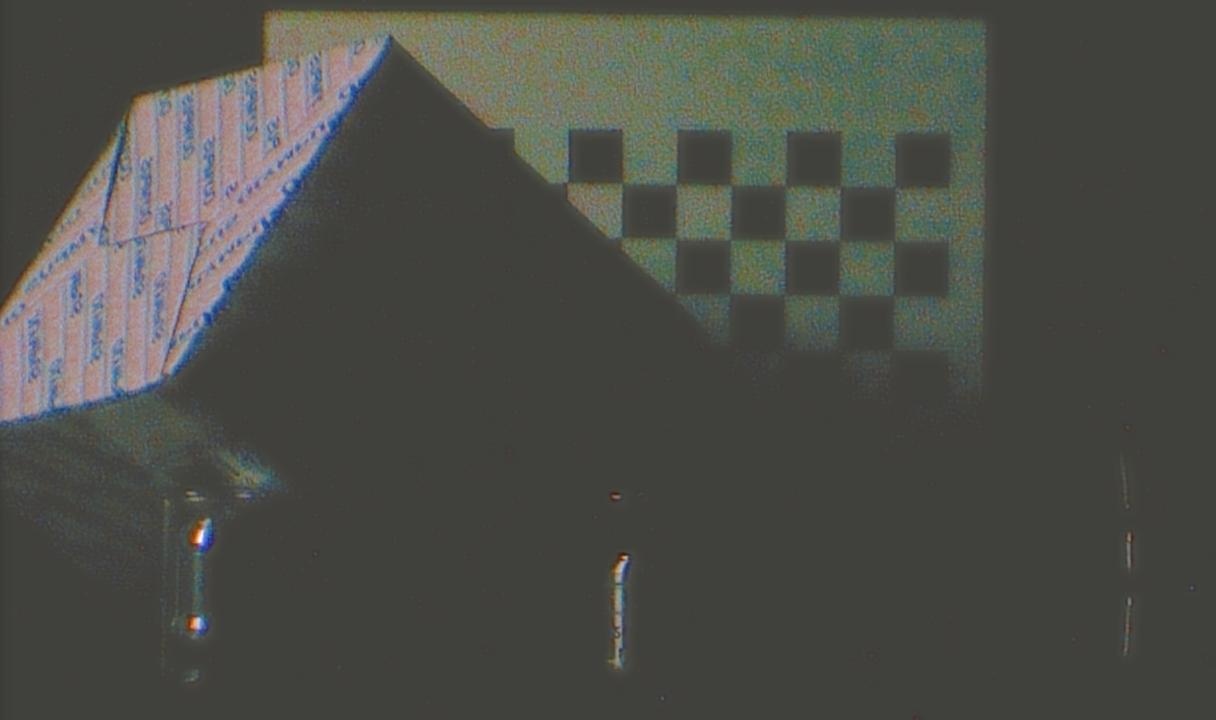}
            \caption*{RetinexNet~\cite{wei2018deep}}
        \end{subfigure}
        \begin{subfigure}{0.195\textwidth}
            \includegraphics[width=\linewidth]{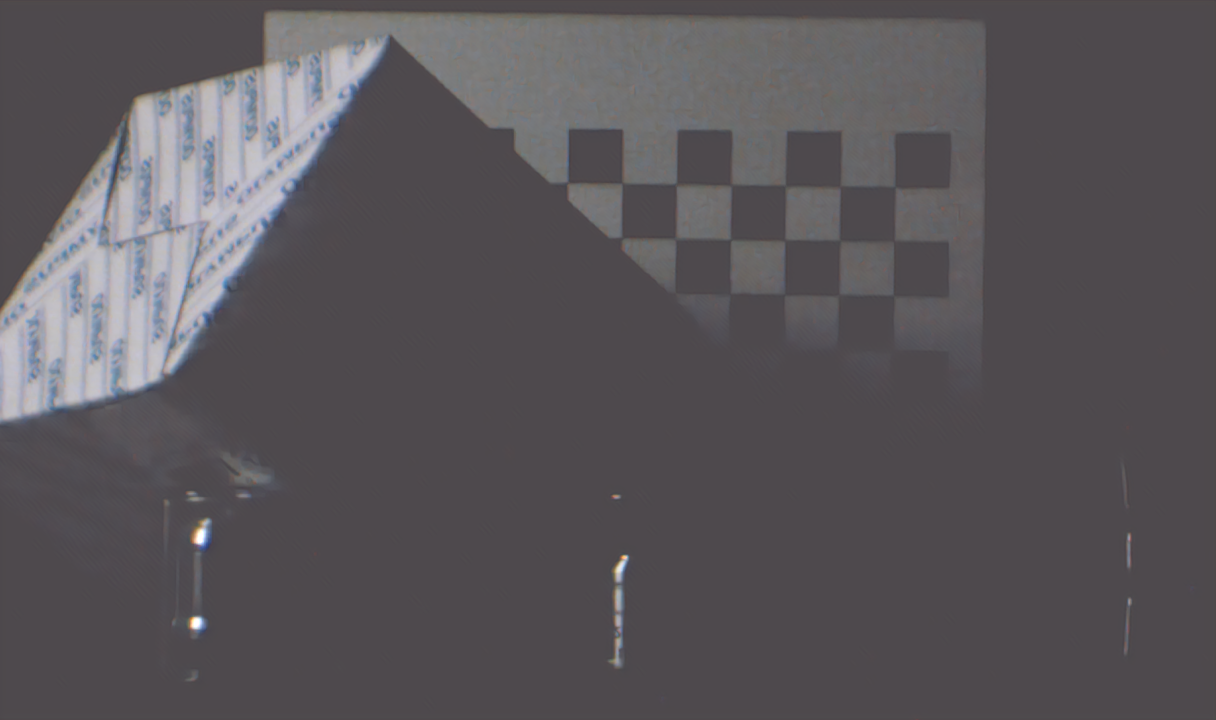}
            \caption*{URetinex-Net~\cite{wu2022uretinex5}}
        \end{subfigure}
        \begin{subfigure}{0.195\textwidth}
            \includegraphics[width=\linewidth]{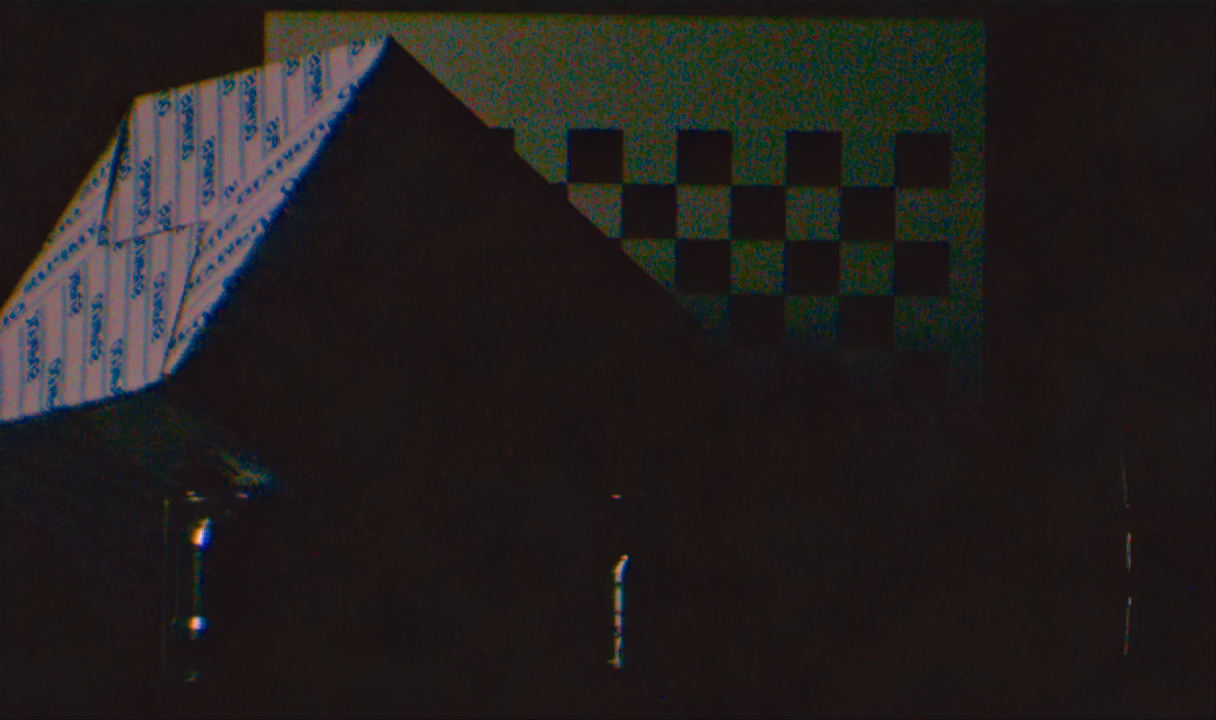}
            \caption*{GSAD~\cite{hou2024global}}
        \end{subfigure}
        \begin{subfigure}{0.195\textwidth}
            \includegraphics[width=\linewidth]{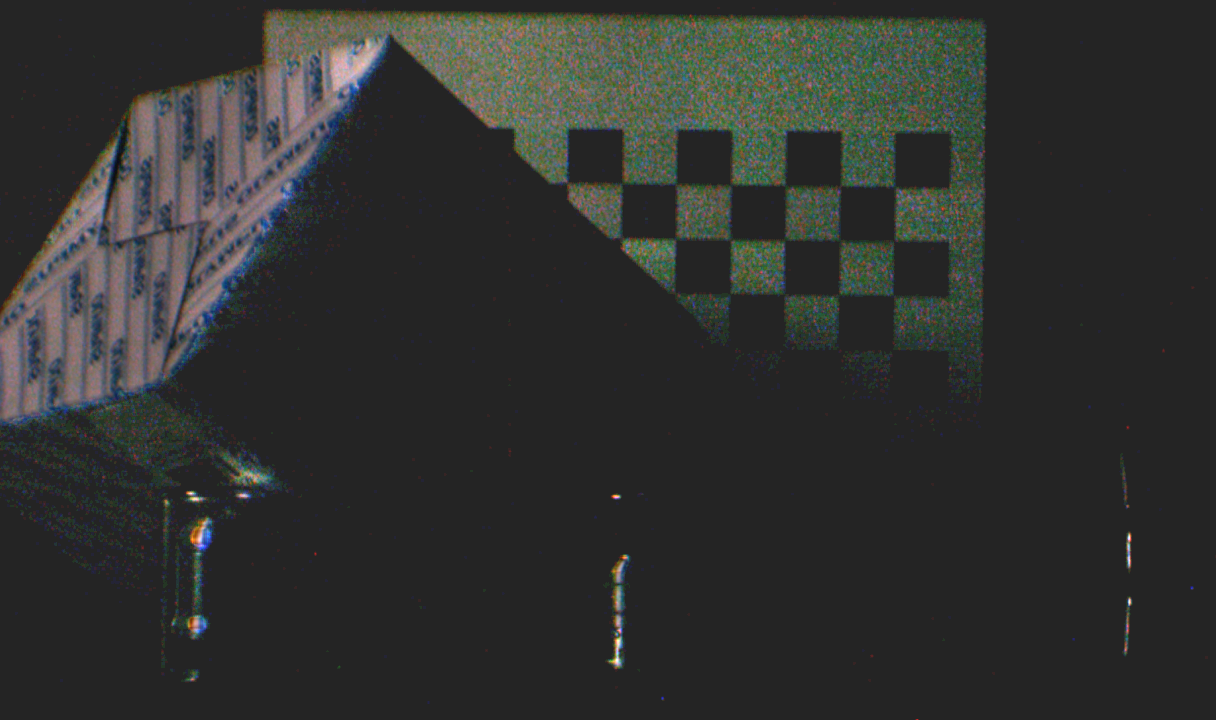}
            \caption*{LIME~\cite{guo2016lime}}
        \end{subfigure}
        \vspace{-2pt}
    
        \begin{subfigure}{0.195\textwidth}
            \includegraphics[width=\linewidth]{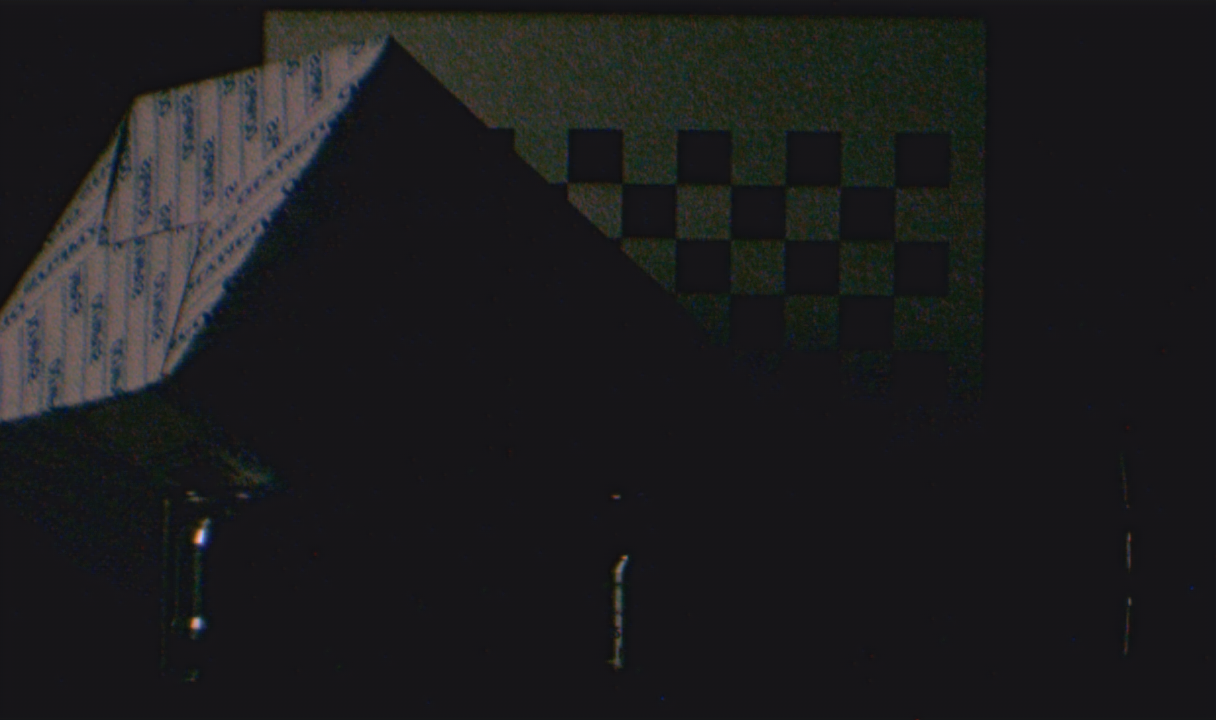}
            \caption*{Retinexformer~\cite{cai2023retinexformer}}
        \end{subfigure}
        \begin{subfigure}{0.195\textwidth}
            \includegraphics[width=\linewidth]{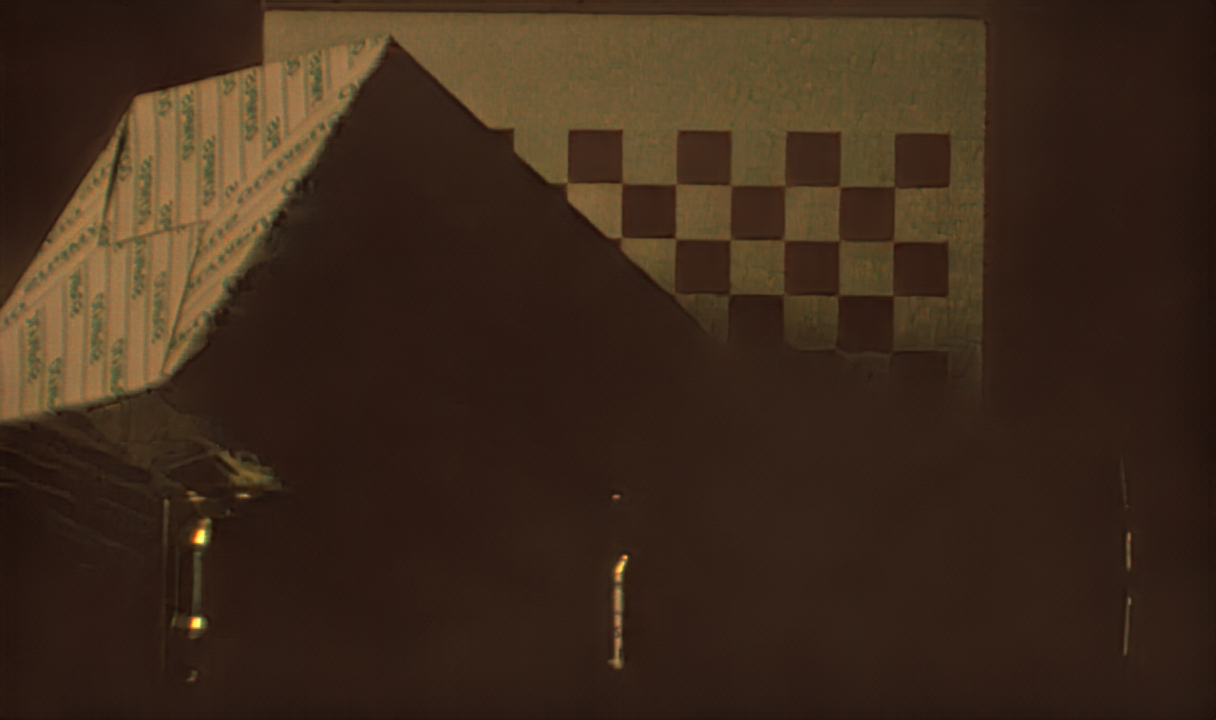}
            \caption*{ELIE~\cite{jiang2023event}}
        \end{subfigure}
        \begin{subfigure}{0.195\textwidth}
            \includegraphics[width=\linewidth]{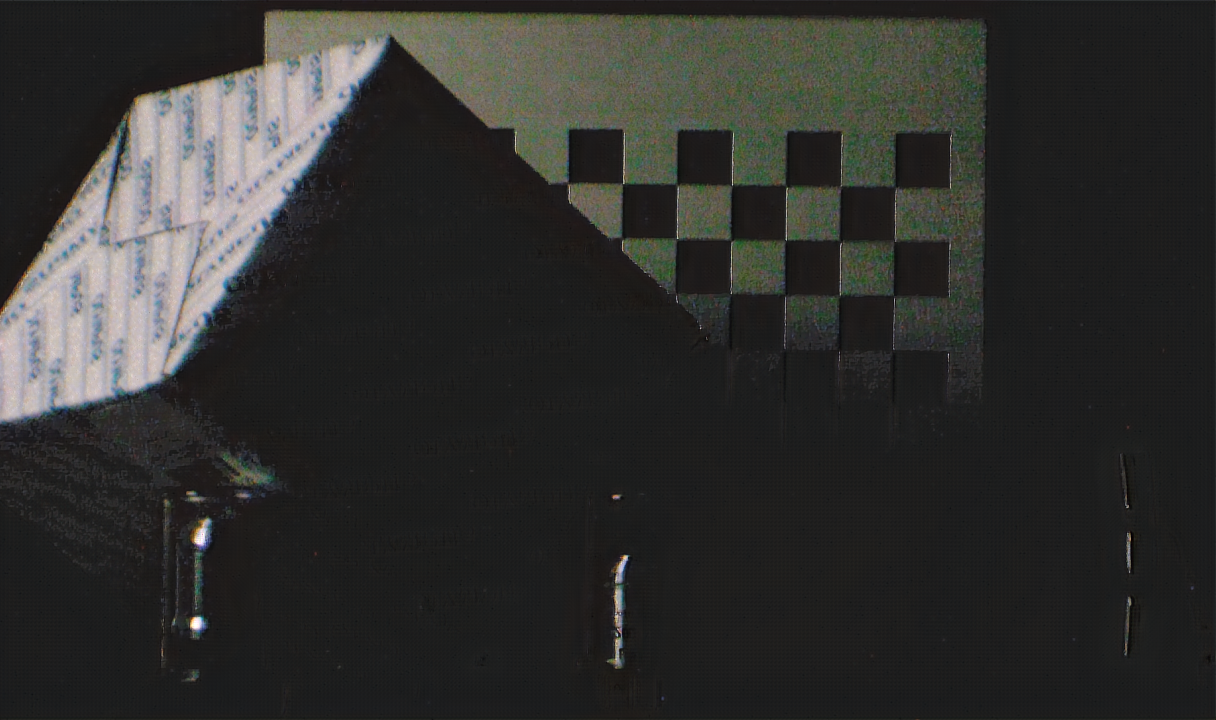}
            \caption*{EvLight~\cite{liang2024towards}}
        \end{subfigure}
        \begin{subfigure}{0.195\textwidth}
            \includegraphics[width=\linewidth]{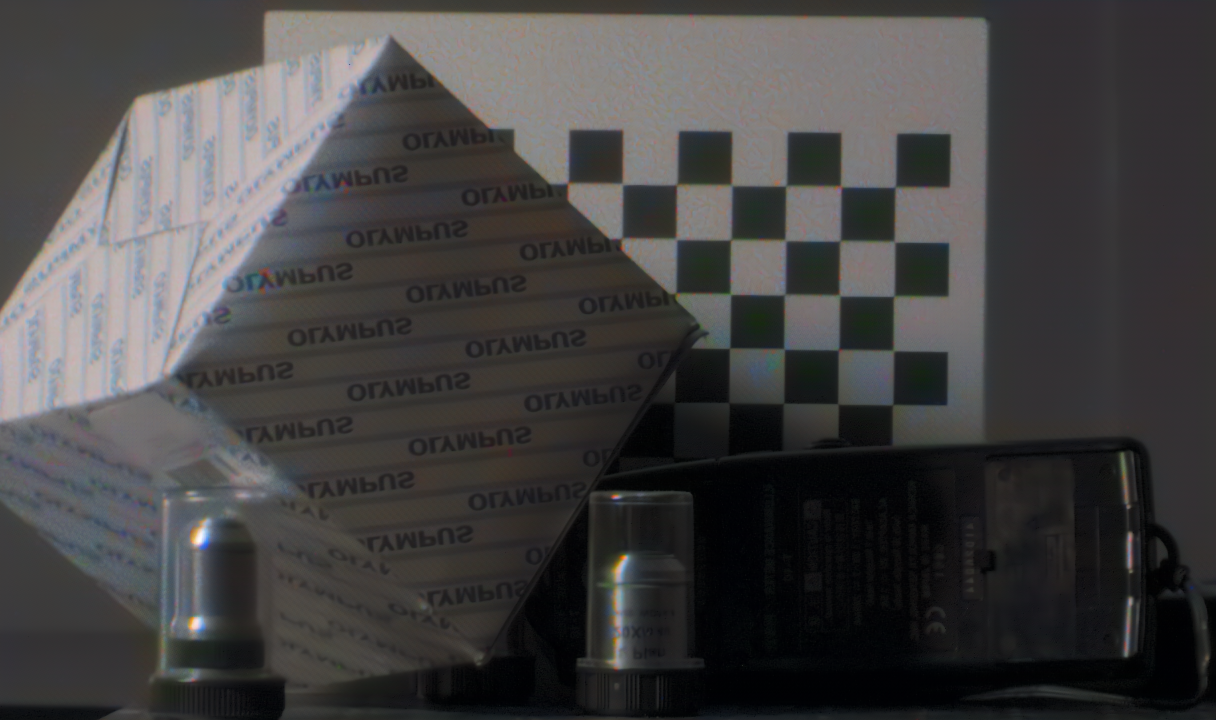}
            \caption*{\textbf{\ourmethod (Ours)}}
        \end{subfigure}
        \begin{subfigure}{0.195\textwidth}
            \includegraphics[width=\linewidth]{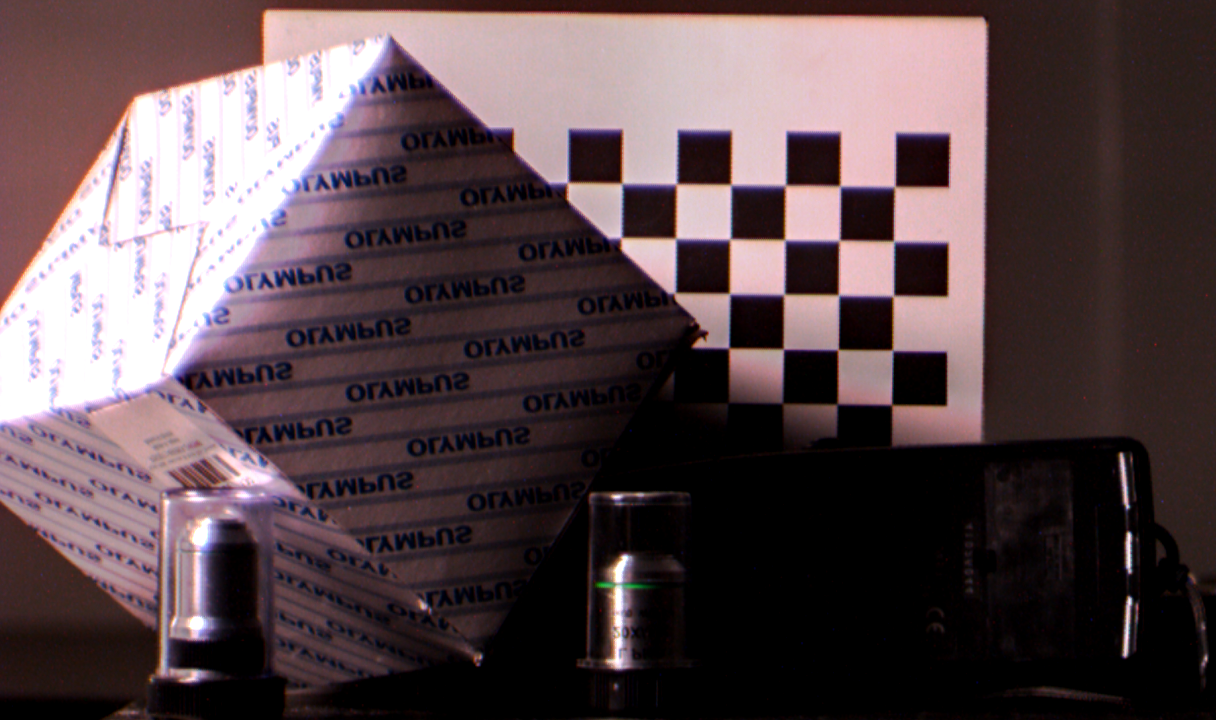}
            \caption*{GT}
        \end{subfigure}
    
        \setcounter{subfigure}{0} 
        \renewcommand{\name}{figures/qualitative_ourdataset/41low1/}
        \begin{subfigure}{0.195\textwidth}
            \includegraphics[width=\linewidth]{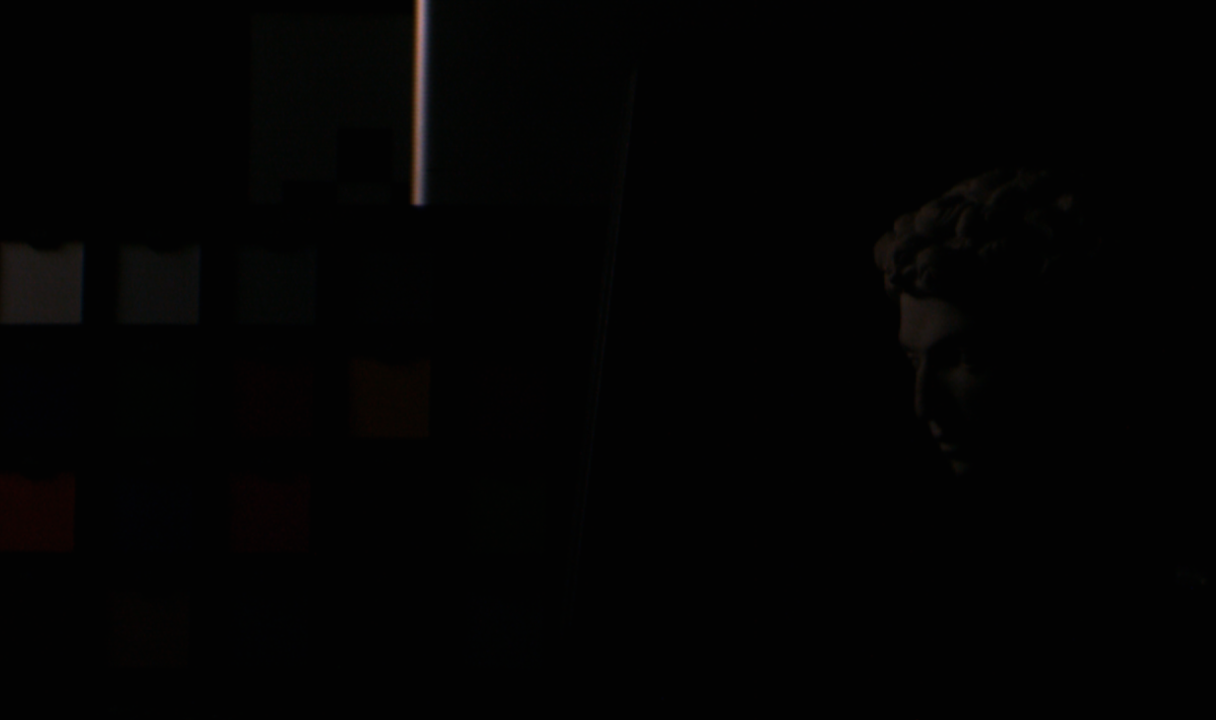}
            \caption*{Input}
        \end{subfigure}
        \begin{subfigure}{0.195\textwidth}
            \includegraphics[width=\linewidth]{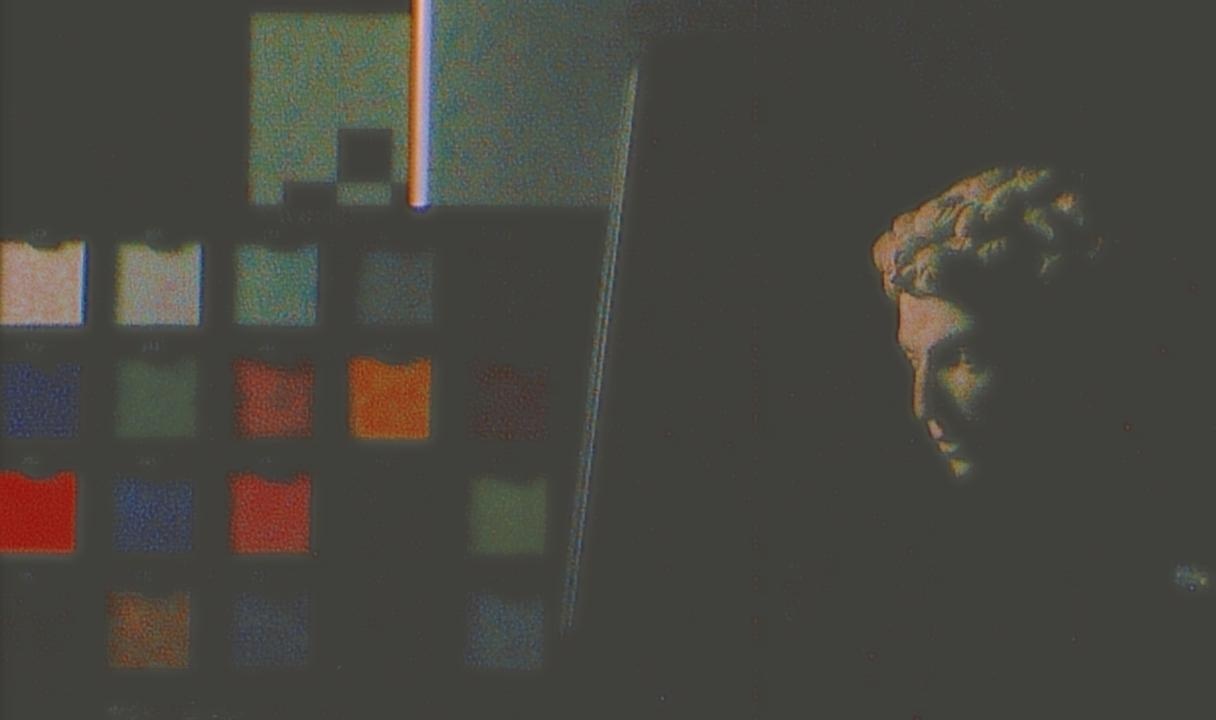}
            \caption*{RetinexNet~\cite{wei2018deep}}
        \end{subfigure}
        \begin{subfigure}{0.195\textwidth}
            \includegraphics[width=\linewidth]{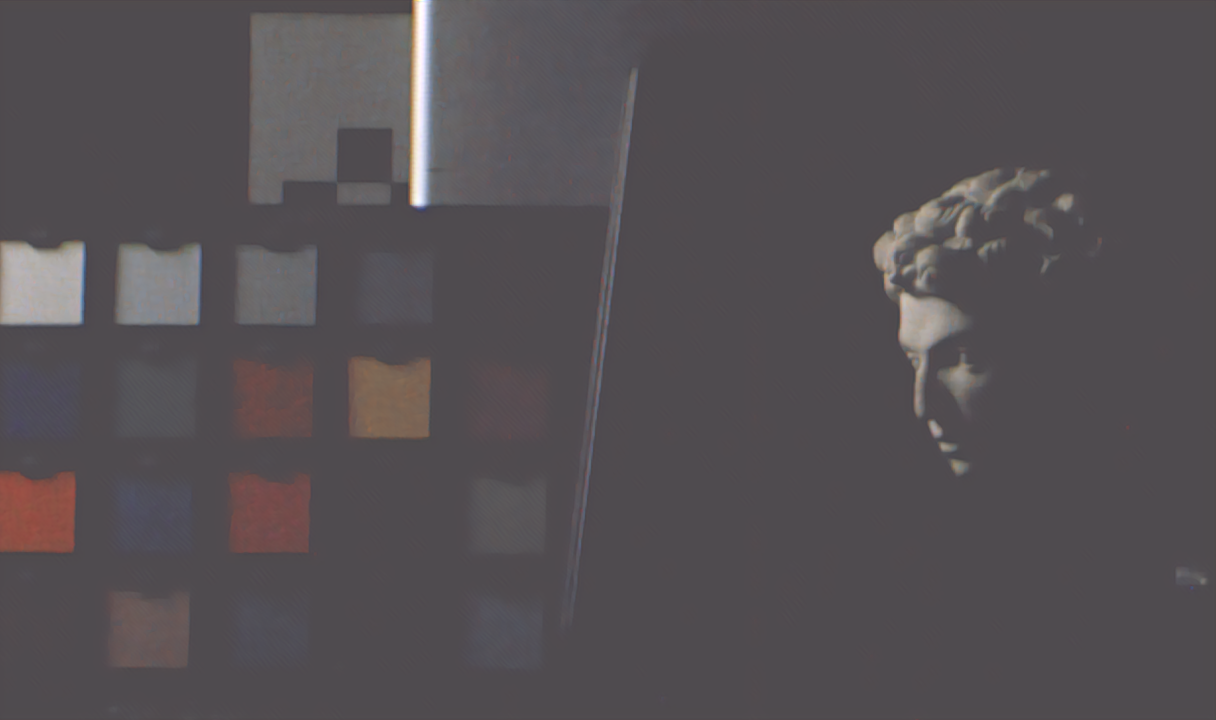}
            \caption*{URetinex-Net~\cite{wu2022uretinex5}}
        \end{subfigure}
        \begin{subfigure}{0.195\textwidth}
            \includegraphics[width=\linewidth]{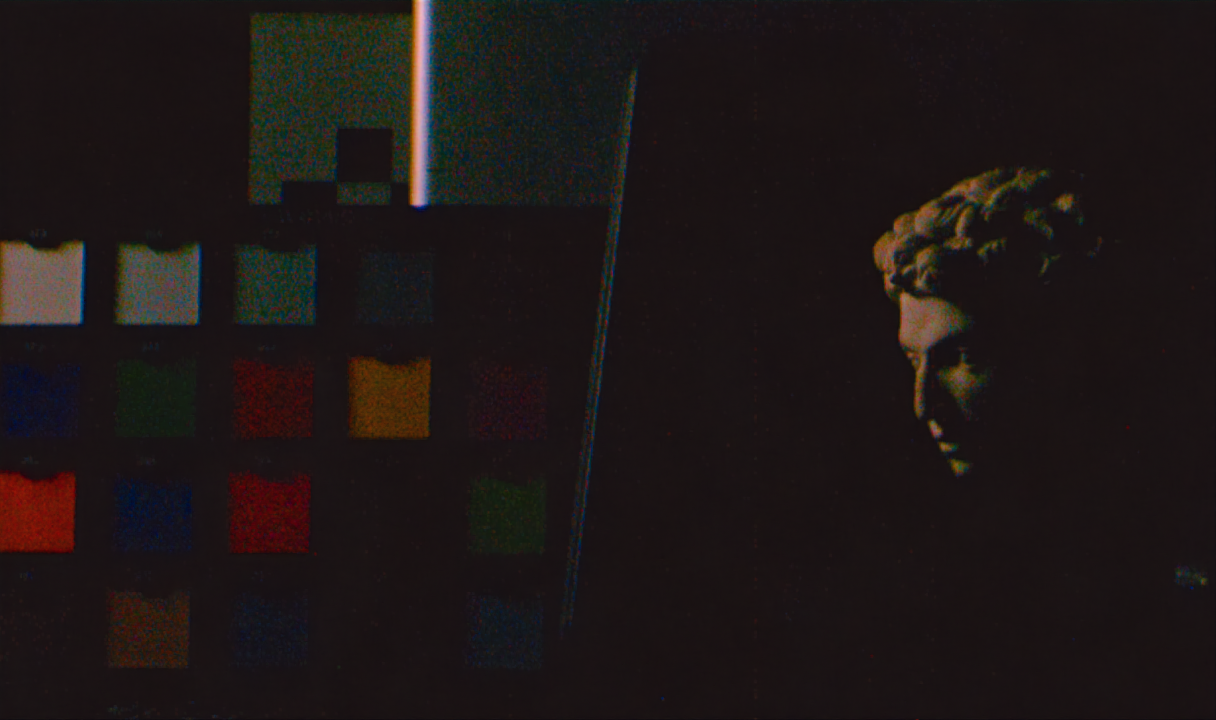}
            \caption*{GSAD~\cite{hou2024global}}
        \end{subfigure}
        \begin{subfigure}{0.195\textwidth}
            \includegraphics[width=\linewidth]{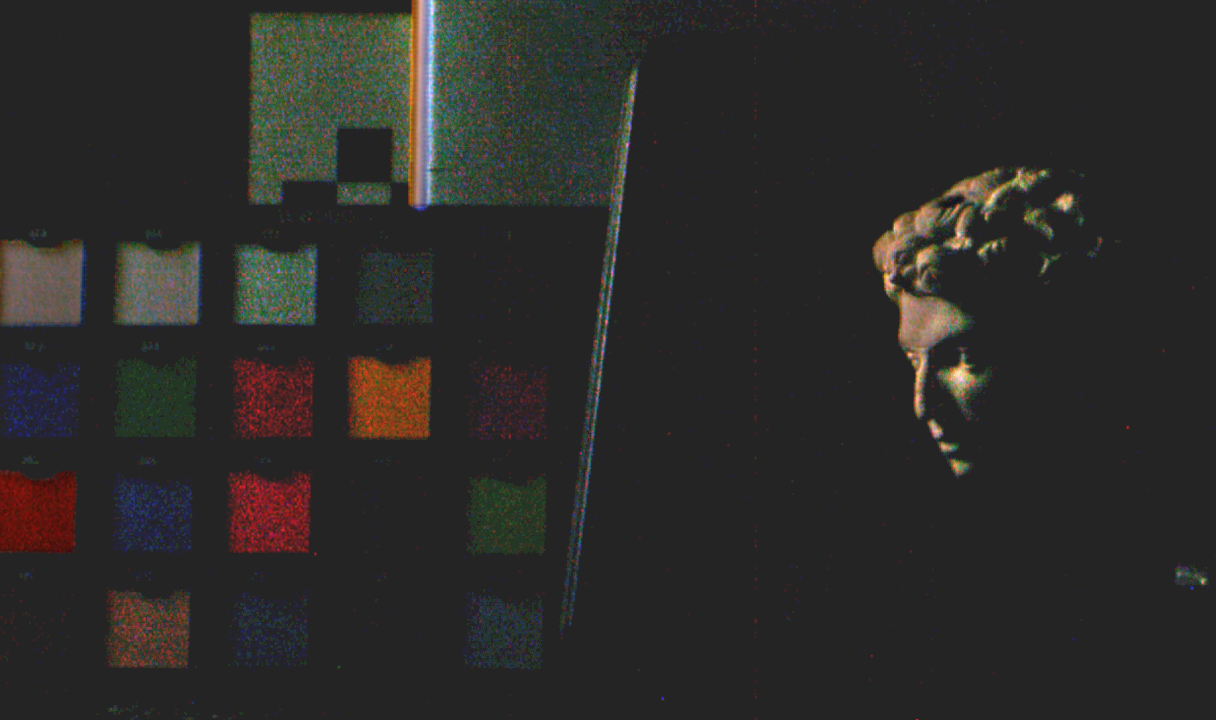}
            \caption*{LIME~\cite{guo2016lime}}
        \end{subfigure}
        \vspace{-2pt}
    
        \begin{subfigure}{0.195\textwidth}
            \includegraphics[width=\linewidth]{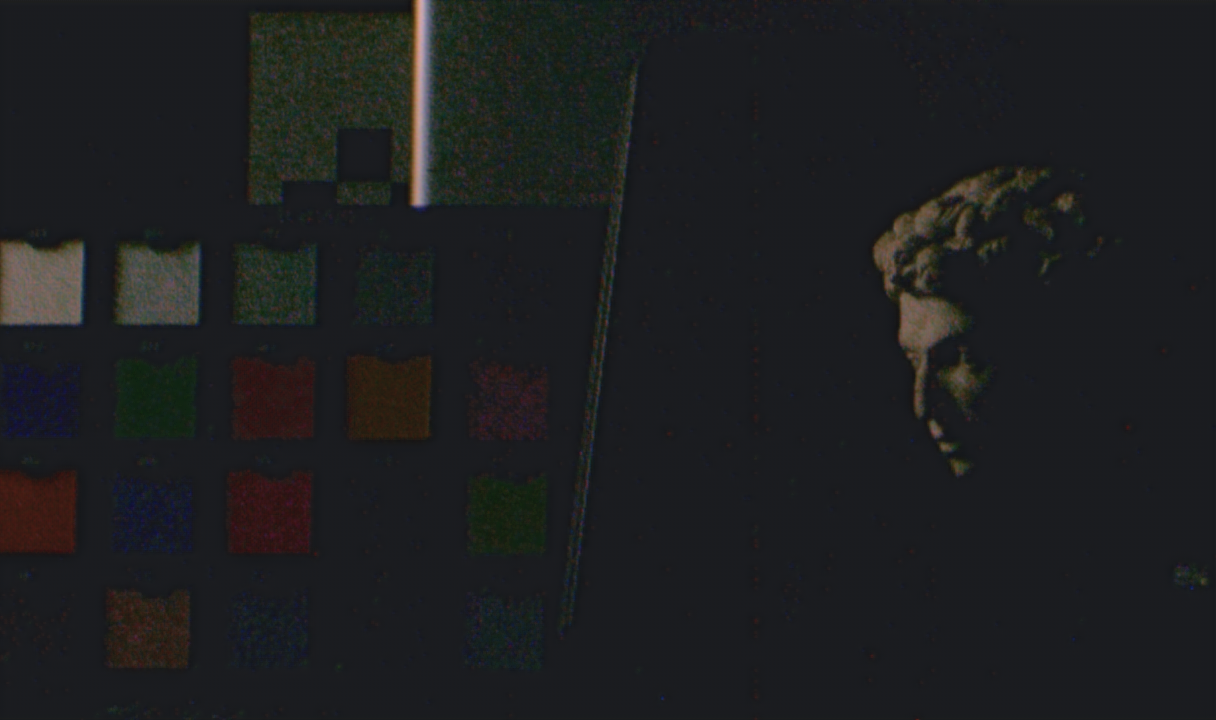}
            \caption*{Retinexformer~\cite{cai2023retinexformer}}
        \end{subfigure}
        \begin{subfigure}{0.195\textwidth}
            \includegraphics[width=\linewidth]{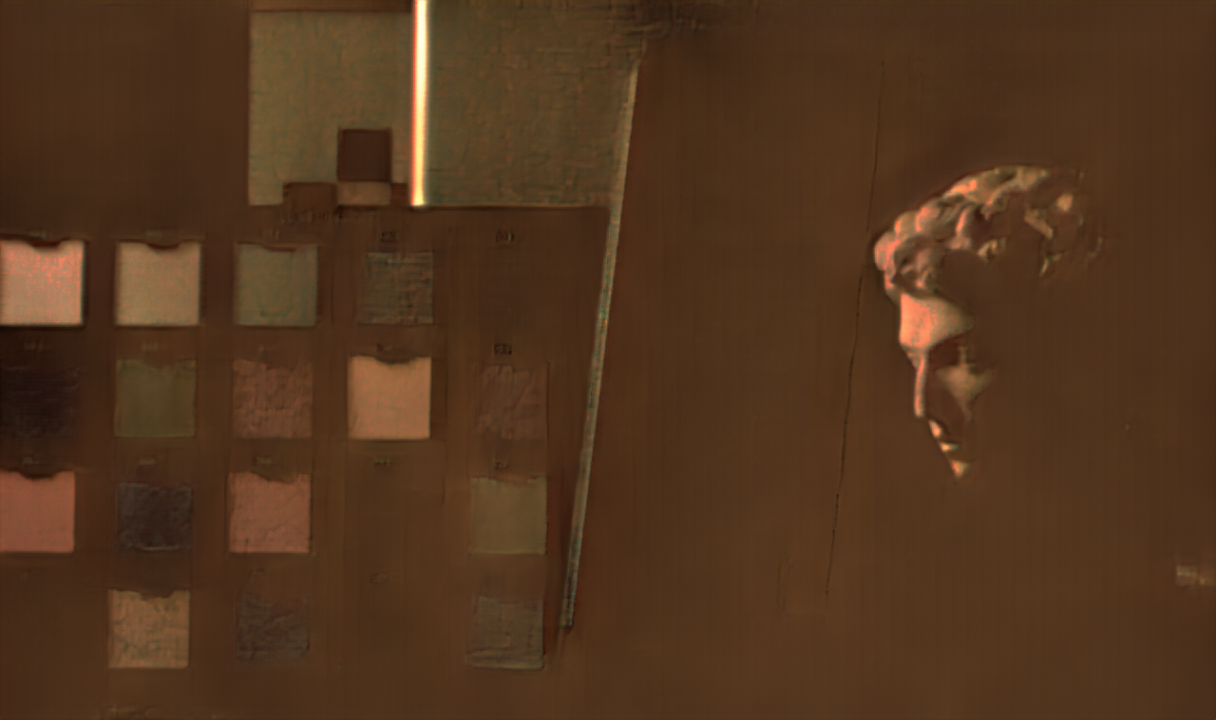}
            \caption*{ELIE~\cite{jiang2023event}}
        \end{subfigure}
        \begin{subfigure}{0.195\textwidth}
            \includegraphics[width=\linewidth]{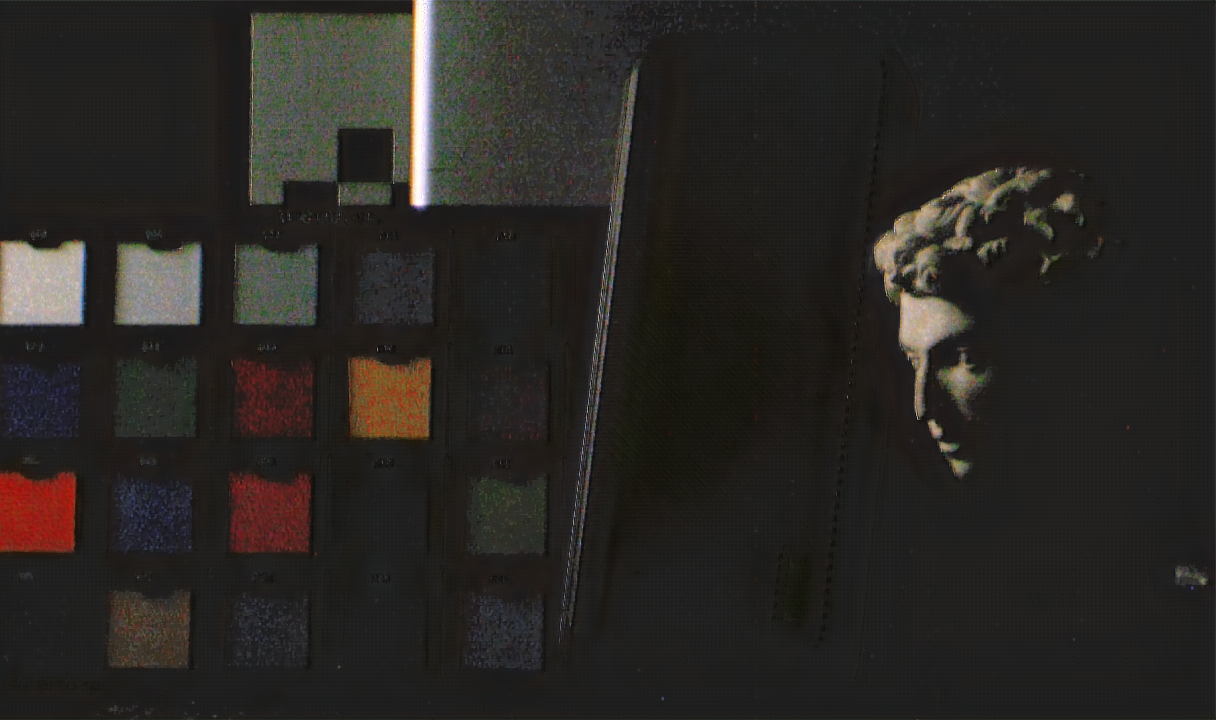}
            \caption*{EvLight~\cite{liang2024towards}}
        \end{subfigure}
        \begin{subfigure}{0.195\textwidth}
            \includegraphics[width=\linewidth]{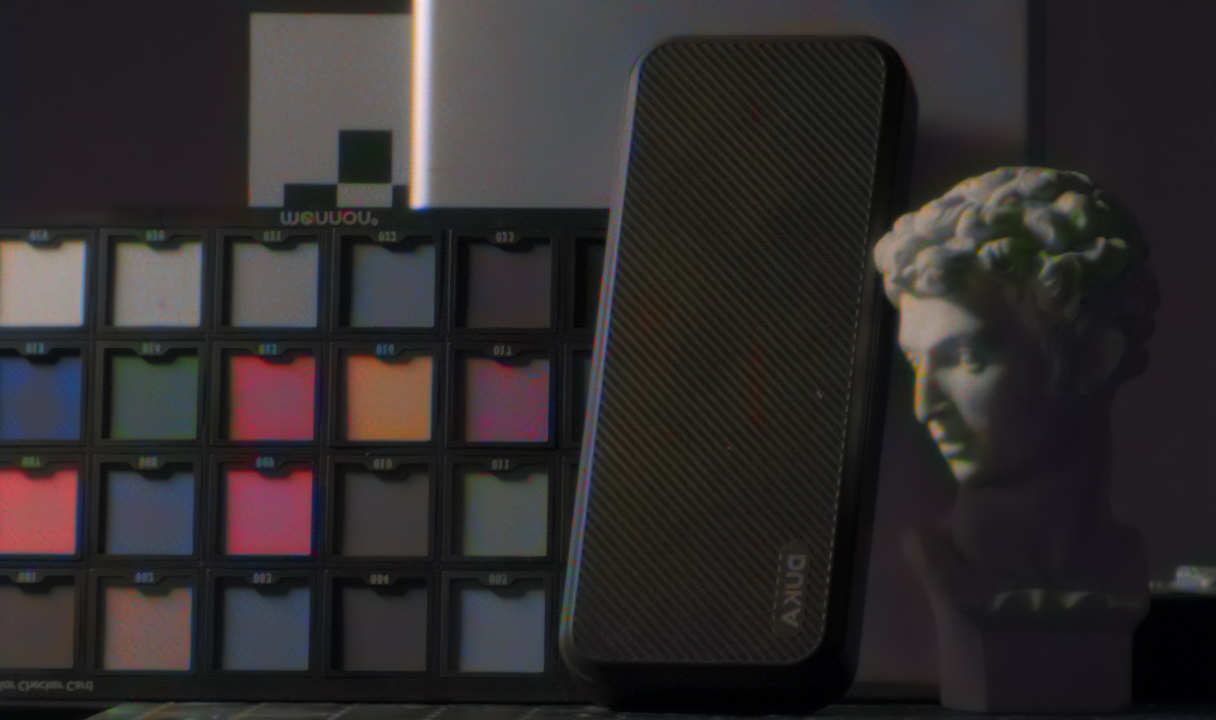}
            \caption*{\textbf{\ourmethod (Ours)}}
        \end{subfigure}
        \begin{subfigure}{0.195\textwidth}
            \includegraphics[width=\linewidth]{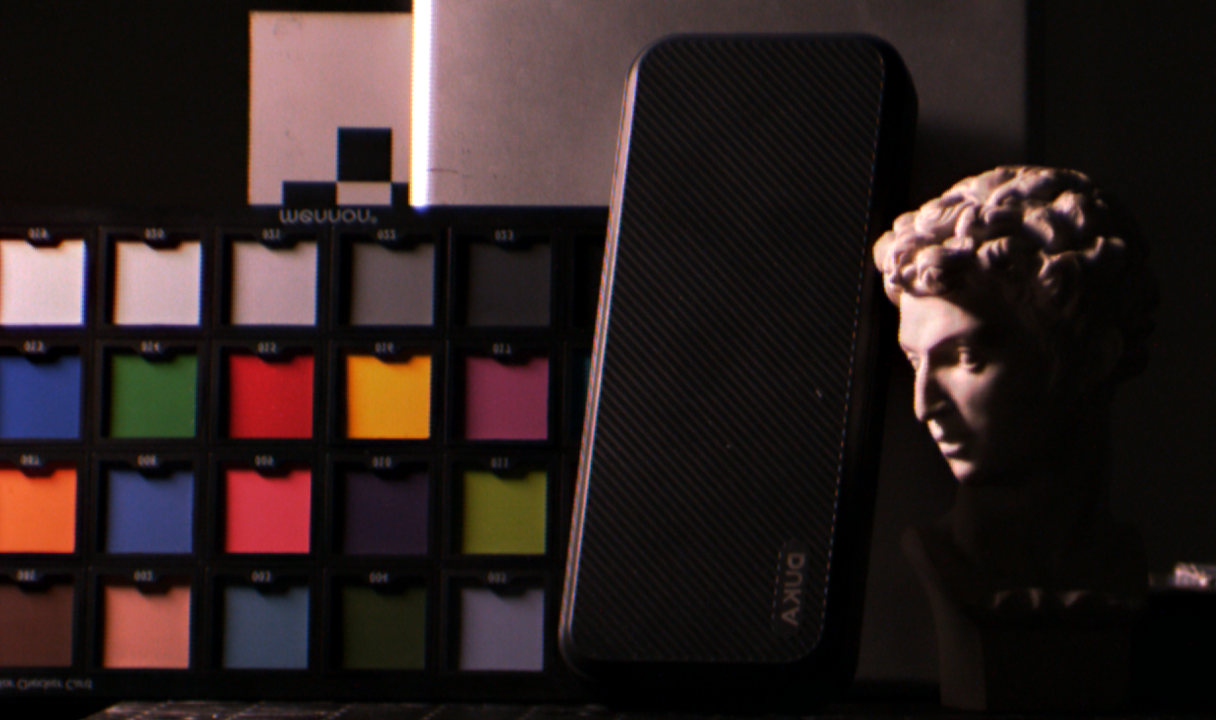}
            \caption*{GT}
        \end{subfigure}
    
        \vspace{-8pt}
        \caption{\textbf{Visual comparison with state-of-the-art methods on \ourdataset}. \ourmethod improves visibility and HDR performance.}
        \vspace{-10pt}
        \label{fig:qualitative_ours}
    \end{figure*}

    \subsection{Datasets}
    \label{subsec:settings}
    
    \PAR{LOL v1 \& v2.} The images in the LOL v1~\cite{wei2018deep} and LOL v2 Real~\cite{yang2020fidelity} datasets have a resolution of $600 \times 400$, whereas the images in the LOL v2 Synthetic~\cite{yang2020fidelity} dataset have a resolution of $384 \times 284$. The training and testing sets for LOL v1, LOL v2 Real, and LOL v2 Synthetic datasets are divided in the ratios of 485:15, 689:100, and 900:100, resp.

    
    \PAR{SDSD.} The SDSD dataset~\cite{wang2021seeing} comprises both indoor and outdoor subsets. We utilize 70:12 video pairs for training and testing in the indoor subset and 80:13 video pairs for the outdoor subset. All images have a resolution of $512\times960$.

    \PAR{\ourdataset.} 60 groups of data are present in \ourdataset. Each set consists of 3 low-light images, 1 HDR image, temporal-mapping events, and motion events.

    \subsection{Training Details}
    \label{subsec:training_details}
    
    

    We synthesize events with contrast threshold $c$ set randomly following Gaussian distribution $\mathcal{N}(\mu=0.2, \sigma=0.03)$ and $\mathcal{N}(\mu=0.2, \sigma=0.05)$ during training and testing stage to simulate scenarios encountered during testing with different camera settings. 
    The network is trained with a patch size of $128\times128$. The denoising network in T2I module is pre-trained for 100k iterations with the proposed degradation model and geometric augmentation (random flips and rotations) for data augmentation. Learning rate is set to $2\times10^{-4}$ with Adam~\cite{kingma2014adam} optimizer and minimum learning rate of $10^{-7}$. The learning rate is the same as pre-training except for the denoising network, which is set to one-tenth of that for the other components. We train the model with a batch size of 32 for 150k iterations on a single NVIDIA L4 GPU. FPS are evaluated with a single NVIDIA A100 GPU with $640\times480$ images. For more details kindly \cf supp.

    \subsection{Comparisons on Synthetic Datasets}
    
    We compare our method with state-of-the-art techniques on five synthetic datasets. PSNR results are computed without using the mean value from ground truth~\cite{wu2023learning,zhang2019kindling13}. Motion-based methods, such as ELIE~\cite{jiang2023event}, Liu~\etal~\cite{liu2023low}, and EvLight~\cite{liang2024towards}, cannot be applied to the LOL v1 and v2 datasets as they require video sequences to generate motion events.

    \PAR{Quantitative Comparison.}~
    The quantitative results are reported in Table~\ref{tab:synthetic}. Compared to the best existing image-only and event-based methods, our \ourmethod achieves 3.44/7.52/6.39/3.88/3.45 dB improvement in PSNR and 0.032/0.089/0.021/0.047/0.081 improvement in SSIM, respectively, with only 3.44 Mb parameters. With events as prior for illumination estimation, our method significantly outperforms image-only methods. As a representative method for event-based methods leveraging motion events for LLIE, EvLight~\cite{liang2024towards} only achieves limited improvements in SSIM (0.017) due to the edge information from motion events. However, EvLight fails to outperform other image-only methods in other dimensions even with the event as an additional modality. By utilizing temporal-mapping events, our method fully harnesses the potential of event-based information with a state-of-the-art performance among event-based methods.
    
    \PAR{Visual Comparison.}
    Fig.~\ref{fig:qualitative_lolv1} shows example results from LIME~\cite{guo2016lime}, URetinex-Net\cite{wu2022uretinex5}, GSAD~\cite{hou2024global}, Retinexformer~\cite{cai2023retinexformer} and our method in LOL v1 dataset. Our method restores the best details in the poor-illumination area, \eg, seats in the dark in the red box.

    \begin{table*}[!t]
    \centering
    \small
    \setlength{\tabcolsep}{10pt}
    \renewcommand\arraystretch{1.1}
    \caption{\textbf{Quantitative comparison and user study on \ourdataset,} with the best results highlighted in \textbf{bold}. PSNR$^{*}$ and SSIM$^{*}$ are calculated with exposure fusion images. Avg ranking: the average ranking from the user study. FPS: Frame-per-second.} 
    \vspace{-8pt}
    \resizebox{0.98\textwidth}{!}{%
        \begin{tabular}{r||cccccccc}
        \bottomrule[0.12em]
        \rowcolor{tableHeadGray}
        \textbf{Method} &  \textbf{Events} & \textbf{PSNR* $\uparrow$}  & \textbf{SSIM* $\uparrow$}  &  \textbf{PIQE $\downarrow$}  & \textbf{Avg Ranking $\downarrow$}  & \textbf{Param (M) $\downarrow$} & \textbf{Flops (G)$\downarrow$} & \textbf{FPS $\uparrow$} \\ \hline \hline
        LIME~\pub{TIP2016}~\cite{guo2016lime}  & \xmark    &  13.31 & 0.400 & 41.68 & 2.34 & - & \textbf{7.0} & 0.77  \\ 
        RetinexNet~\pub{BMVC2018}~\cite{wei2018deep} &  \xmark    &  12.66 & 0.363  & 15.31 & 4.31 & 0.84 & 246.4 & 14.4 \\ 
        URetinex-Net~\pub{CVPR2022}~\cite{wu2022uretinex5} & \xmark    &  12.29 & 0.382 & 45.97 & 3.93 & \textbf{0.34} & 106.5 & \textbf{48.1} \\ 
        Retinexformer~\pub{ICCV2023}~\cite{cai2023retinexformer}  & \xmark    &  11.79 & 0.403  & 19.55 & 5.93 & 1.61 & 235.1 & 3.7 \\ 
        GSAD~\pub{NeuralIPS2024}~\cite{hou2024global}&  \xmark    &  12.20 & 0.416 & 31.23 & 3.54 & 17.36 & 942.6 & 1.6 \\ 
    \arrayrulecolor{gray}\hdashline\arrayrulecolor{black} 
        ELIE~\pub{TMM2023}~\cite{yu2023learning} &  \cmark    &  13.67 & 0.424 & 42.73 & 3.57 & 204.95 & 2116.0 & 0.9 \\ 
        EvLight~\pub{CVPR2024}~\cite{liang2024towards} &  \cmark    &  14.51 & 0.435 & 20.45 & 2.31 & 22.73 & 438.5 & 4.1 \\ 
        \textbf{\ourmethod (Ours)} &  \cmark  &    \textbf{15.39} & \textbf{0.470} & \textbf{9.41} & \textbf{1.21} & 3.44 & 184.6 & 35.6 \\ 
        \hline
    
        \end{tabular}}
        \vspace{-15pt}
    
        \label{tab:realworld}
    \end{table*}

    \begin{table}[!t]
        \caption{\textbf{Ablation study of various components of our method} on LOL v1 dataset~\cite{wei2018deep}. ``IRE'': Illumination-aid Reflectance Enhancement module, ``LLDM'': Low-light degradation model. ``CrossAttn'': Cross-modal attention.}
        \vspace{-10pt}
        \resizebox{0.48\textwidth}{!}{%
        \setlength{\tabcolsep}{12pt}
        \renewcommand\arraystretch{1.1}
        \begin{tabular}{cccccc}
        \bottomrule[0.12em]
        \rowcolor{tableHeadGray}
          & \textbf{Events} & \textbf{T2I module}      & \textbf{IRE} & PSNR & SSIM  \\ \hline \hline
         1 & \xmark  & {n/a}                     & n/a  & 16.98 & 0.565 \\
         2 & \cmark  & None                     & n/a  & 26.96 & 0.849 \\
         3 & \cmark  & \textit{w./o.} LLDM      & No fusion & 27.25 & 0.862 \\
         4 & \cmark  & \textit{w./o.} LLDM      & CrossAttn. fusion & 28.45 & 0.872 \\
         5 & \cmark  & \textit{w.} LLDM         & No fusion  & 27.83  & 0.865 \\
          6 & \cmark & \textit{w.} LLDM         & Add fusion  & 28.13 & 0.870 \\
         7 & \cmark & \textit{w.} LLDM         & Concat. fusion  & 28.01 & 0.867 \\
         8 & \cmark & \textit{w.} LLDM         & Multiply fusion  & 27.98 & 0.867 \\
         9 & \cmark  & \textit{w.} LLDM         & CrossAttn. fusion  & 28.60 & 0.877 \\
         \hline
        \end{tabular}
        }
        \label{tab:ablation}
        \vspace{-18pt}
    \end{table}

    \subsection{Comparisons on \ourdataset}
    To assess methods in real-world settings, we evaluate state-of-the-art image-only and event-based approaches, including ELIE~\cite{jiang2023event}, EvLight~\cite{liang2024towards}, and our method on \ourdataset. Except for EvLight, trained on SDE-in~\cite{liang2024towards}, all methods are trained on LOL v1 and directly tested on \ourdataset, demonstrating their generalization to real-world scenes.

    \PAR{Visual Comparison.}~As shown in Fig.~\ref{fig:qualitative_ours}, image-only methods show limited performance in the dark area. Other than improving the average brightness of the image, they fail to enhance the contrast in the poor illumination area. Benefiting from motion events, ELIE and EvLight improve contrast better than image-only methods. However, it also suffers from edge artifacts caused by the events, \eg, ghost artifacts around the camera in the second example. Our method demonstrates balanced illumination, making details in dark areas, such as the black box and dark side of the sculpture, clearly visible. Even the gray background with uniformly distributed illumination is well-defined. With the estimated illumination from the T2I module, the enhanced reflectance yields \textit{smoother results with reduced noise}.
    Other than low-light performance, our results also benefit from the advantage of HDR from event cameras. In high dynamic range scenes, the ``Olympus'' logo is clearly visible in both illuminated and backlit areas. 
    In addition to low-light performance, our results \textit{also demonstrate HDR capability}, benefiting from the inherent advantages of event cameras. In high dynamic range scenes, the ‘Olympus’ logo remains clearly visible in both illuminated and backlit areas.

    \PAR{Quantitative Comparison.}~Table~\ref{tab:realworld} shows quantitative results. Note that for PSNR and SSIM calculation, the ground truth image is generated from three single-exposure images using exposure fusion based on Mertens~\etal~\cite{mertens2007exposure}.
    Because of the inherent difference in dynamic range and response function between the frame-based camera and the event camera in our setup, the pixel-wise comparison may not be meaningful~\cite{hartley2003multiple,kim2012new}, yet we report PSNR and SSIM (marked with *) as a reference. Our \ourmethod outperforms all methods in PSNR, SSIM, and the non-reference metric Perception-based Image Quality Evaluator (PIQE)~\cite{venkatanath2015blind}.
    
    \PAR{User Study.} To quantify the human subjective visual perception quality of the enhanced images, we conduct a user study among 43 human subjects. 8 sets of results are randomly selected for the test.
    Testers are invited to rank the results from 1 (best) to 8 (worst). 
    For each low-light image, we present the image alongside results enhanced by various algorithms, without displaying the algorithms' names, to the human testers.
    Results from \ourmethod are most favored by the testers, followed by EvLight, and LIME.

    \subsection{Ablation Study}
    We conduct an ablation study on the LOL v1 dataset to verify the contributions of the proposed components and data augmentation methods, as shown in Table~\ref{tab:ablation}.
    The introduction of temporal-mapping events brings an improvement of 10.51 dB improvement in PSNR, which proves the huge potential of temporal-mapping events in LLIE. Adding the T2I module (row 3) improves the performance by 0.29 dB. 
    The design of the low-light degradation model in the data augmentation of denoising network training brings an improvement of 0.58 dB improvement in PSNR, evidencing the benefit of the modeling of event camera response under low-illumination (rows 3 and 5).
    Introducing illumination in the reflectance enhancement module with simple ``add fusion'' yields 0.64 dB improvements (rows 5 and 6). When changing to cross-modal attention, the improvement comes to 0.47/0.59/0.62 dB compared to ``add'', ``concatenation'', and ``multiply'' fusion.
    Finally, all the components and model designs yield a substantial improvement of 11.62 dB, setting a new state-of-the-art in event-based LLIE. Qualitative results for ablation study \cf supp.


    \section{Conclusion}
    \label{sec:conclusion}
    
    
    In this work, we approach low-light image enhancement through event-based illumination estimation. Unlike previous LLIE methods that extract edge information from motion events, we exploit the event camera’s superior low-light responsiveness by capturing and processing temporal-mapping events for illumination estimation. This well-estimated illumination further improves reflectance decomposition and refinement. Based on Retinex theory, the enhanced image is reconstructed by multiplying reflectance and illumination. We create a real-world dataset using a beam-splitter dual-sensor system. On five synthetic datasets and our \ourdataset, \ourmethod surpasses state-of-the-art methods, achieving real-time inference at 35.6 FPS.

{
\bibliographystyle{ieeenat_fullname} 
\small
\bibliography{arxiv}

\begin{thebibliography}{57}
\providecommand{\natexlab}[1]{#1}
\providecommand{\url}[1]{\texttt{#1}}
\expandafter\ifx\csname urlstyle\endcsname\relax
  \providecommand{\doi}[1]{doi: #1}\else
  \providecommand{\doi}{doi: \begingroup \urlstyle{rm}\Url}\fi

\bibitem[Arici et~al.(2009)Arici, Dikbas, and Altunbasak]{arici2009histogram1}
Tarik Arici, Salih Dikbas, and Yucel Altunbasak.
\newblock A histogram modification framework and its application for image contrast enhancement.
\newblock \emph{IEEE Transactions on image processing}, 18\penalty0 (9):\penalty0 1921--1935, 2009.

\bibitem[Bao et~al.(2024)Bao, Sun, Ma, and Wang]{bao2024temporal}
Yuhan Bao, Lei Sun, Yuqin Ma, and Kaiwei Wang.
\newblock Temporal-mapping photography for event cameras.
\newblock \emph{arXiv preprint arXiv:2403.06443}, 2024.

\bibitem[Brandli et~al.(2014)Brandli, Berner, Yang, Liu, and Delbruck]{brandli2014240}
Christian Brandli, Raphael Berner, Minhao Yang, Shih-Chii Liu, and Tobi Delbruck.
\newblock A 240$\times$ 180 130 db 3 $\mu$s latency global shutter spatiotemporal vision sensor.
\newblock \emph{IEEE Journal of Solid-State Circuits}, 49\penalty0 (10):\penalty0 2333--2341, 2014.

\bibitem[Cai et~al.(2023)Cai, Bian, Lin, Wang, Timofte, and Zhang]{cai2023retinexformer}
Yuanhao Cai, Hao Bian, Jing Lin, Haoqian Wang, Radu Timofte, and Yulun Zhang.
\newblock Retinexformer: One-stage retinex-based transformer for low-light image enhancement.
\newblock In \emph{Proceedings of the IEEE/CVF International Conference on Computer Vision}, pages 12504--12513, 2023.

\bibitem[Chen et~al.(2020)Chen, Teng, Shi, Wang, and Huang]{haoyu2020learning}
Haoyu Chen, Minggui Teng, Boxin Shi, Yizhou Wang, and Tiejun Huang.
\newblock Learning to deblur and generate high frame rate video with an event camera.
\newblock \emph{arXiv preprint arXiv:2003.00847}, 2020.

\bibitem[Debevec et~al.(2004)Debevec, Reinhard, Ward, and Pattanaik]{debevec2004high}
Paul Debevec, Erik Reinhard, Greg Ward, and Sumanta Pattanaik.
\newblock High dynamic range imaging.
\newblock In \emph{ACM SIGGRAPH 2004 Course Notes}, pages 14--es. 2004.

\bibitem[Gallego et~al.(2020)Gallego, Delbr{\"u}ck, Orchard, Bartolozzi, Taba, Censi, Leutenegger, Davison, Conradt, Daniilidis, et~al.]{gallego2020event}
Guillermo Gallego, Tobi Delbr{\"u}ck, Garrick Orchard, Chiara Bartolozzi, Brian Taba, Andrea Censi, Stefan Leutenegger, Andrew~J Davison, J{\"o}rg Conradt, Kostas Daniilidis, et~al.
\newblock Event-based vision: A survey.
\newblock \emph{IEEE Transactions on Pattern Analysis and Machine Intelligence}, 44\penalty0 (1):\penalty0 154--180, 2020.

\bibitem[Guo et~al.(2016)Guo, Li, and Ling]{guo2016lime}
Xiaojie Guo, Yu Li, and Haibin Ling.
\newblock Lime: Low-light image enhancement via illumination map estimation.
\newblock \emph{IEEE Transactions on image processing}, 26\penalty0 (2):\penalty0 982--993, 2016.

\bibitem[Hartley and Zisserman(2003)]{hartley2003multiple}
Richard Hartley and Andrew Zisserman.
\newblock \emph{Multiple view geometry in computer vision}.
\newblock Cambridge university press, 2003.

\bibitem[Hou et~al.(2024)Hou, Zhu, Hou, Liu, Zeng, and Yuan]{hou2024global}
Jinhui Hou, Zhiyu Zhu, Junhui Hou, Hui Liu, Huanqiang Zeng, and Hui Yuan.
\newblock Global structure-aware diffusion process for low-light image enhancement.
\newblock \emph{Advances in Neural Information Processing Systems}, 36, 2024.

\bibitem[iniVation A~G.(2020)]{whitepaper}
iniVation A~G.
\newblock Understanding the performance of neuromorphic event-based vision sensors.
\newblock \emph{Tech. Rep.}, 2020.

\bibitem[Jiang et~al.(2023{\natexlab{a}})Jiang, Luo, Fan, Han, and Liu]{jiang2023low}
Hai Jiang, Ao Luo, Haoqiang Fan, Songchen Han, and Shuaicheng Liu.
\newblock Low-light image enhancement with wavelet-based diffusion models.
\newblock \emph{ACM Transactions on Graphics (TOG)}, 42\penalty0 (6):\penalty0 1--14, 2023{\natexlab{a}}.

\bibitem[Jiang et~al.(2021)Jiang, Gong, Liu, Cheng, Fang, Shen, Yang, Zhou, and Wang]{jiang2021enlightengan}
Yifan Jiang, Xinyu Gong, Ding Liu, Yu Cheng, Chen Fang, Xiaohui Shen, Jianchao Yang, Pan Zhou, and Zhangyang Wang.
\newblock Enlightengan: Deep light enhancement without paired supervision.
\newblock \emph{IEEE transactions on image processing}, 30:\penalty0 2340--2349, 2021.

\bibitem[Jiang et~al.(2023{\natexlab{b}})Jiang, Wang, Li, Zhang, Zhao, and Gao]{jiang2023event}
Yu Jiang, Yuehang Wang, Siqi Li, Yongji Zhang, Minghao Zhao, and Yue Gao.
\newblock Event-based low-illumination image enhancement.
\newblock \emph{IEEE Transactions on Multimedia}, 2023{\natexlab{b}}.

\bibitem[Jobson et~al.(1997)Jobson, Rahman, and Woodell]{jobson1997multiscale}
Daniel~J Jobson, Zia-ur Rahman, and Glenn~A Woodell.
\newblock A multiscale retinex for bridging the gap between color images and the human observation of scenes.
\newblock \emph{IEEE Transactions on Image processing}, 6\penalty0 (7):\penalty0 965--976, 1997.

\bibitem[Kim et~al.(2012)Kim, Lin, Lu, S{\"u}sstrunk, Lin, and Brown]{kim2012new}
Seon~Joo Kim, Hai~Ting Lin, Zheng Lu, Sabine S{\"u}sstrunk, Stephen Lin, and Michael~S Brown.
\newblock A new in-camera imaging model for color computer vision and its application.
\newblock \emph{IEEE Transactions on Pattern Analysis and Machine Intelligence}, 34\penalty0 (12):\penalty0 2289--2302, 2012.

\bibitem[Kim et~al.(2021)Kim, Lee, Wang, and Yoon]{kim2021event}
Taewoo Kim, Jungmin Lee, Lin Wang, and Kuk-Jin Yoon.
\newblock Event-guided deblurring of unknown exposure time videos.
\newblock \emph{arXiv preprint arXiv:2112.06988}, 2021.

\bibitem[Kingma and Ba(2014)]{kingma2014adam}
Diederik~P Kingma and Jimmy Ba.
\newblock Adam: A method for stochastic optimization.
\newblock \emph{arXiv preprint arXiv:1412.6980}, 2014.

\bibitem[Land and McCann(1971)]{land1971retinex}
Edwin~H Land and John~J McCann.
\newblock Lightness and retinex theory.
\newblock \emph{Journal of the Optical Society of America}, 61\penalty0 (1):\penalty0 1--11, 1971.

\bibitem[Li et~al.(2021)Li, Guo, Han, Jiang, Cheng, Gu, and Loy]{li2021low}
Chongyi Li, Chunle Guo, Linghao Han, Jun Jiang, Ming-Ming Cheng, Jinwei Gu, and Chen~Change Loy.
\newblock Low-light image and video enhancement using deep learning: A survey.
\newblock \emph{IEEE transactions on pattern analysis and machine intelligence}, 44\penalty0 (12):\penalty0 9396--9416, 2021.

\bibitem[Liang et~al.(2024)Liang, Chen, Li, Lu, and Wang]{liang2024towards}
Guoqiang Liang, Kanghao Chen, Hangyu Li, Yunfan Lu, and Lin Wang.
\newblock Towards robust event-guided low-light image enhancement: A large-scale real-world event-image dataset and novel approach.
\newblock In \emph{Proceedings of the IEEE/CVF Conference on Computer Vision and Pattern Recognition}, pages 23--33, 2024.

\bibitem[Liang et~al.(2021)Liang, Cao, Sun, Zhang, Van~Gool, and Timofte]{liang2021swinir}
Jingyun Liang, Jiezhang Cao, Guolei Sun, Kai Zhang, Luc Van~Gool, and Radu Timofte.
\newblock Swinir: Image restoration using swin transformer.
\newblock In \emph{Proceedings of the IEEE/CVF international conference on computer vision}, pages 1833--1844, 2021.

\bibitem[Liang et~al.(2023)Liang, Yang, Li, Duan, Xu, and Shi]{liang2023coherent}
Jinxiu Liang, Yixin Yang, Boyu Li, Peiqi Duan, Yong Xu, and Boxin Shi.
\newblock Coherent event guided low-light video enhancement.
\newblock In \emph{Proceedings of the IEEE/CVF International Conference on Computer Vision}, pages 10615--10625, 2023.

\bibitem[Lin et~al.(2020)Lin, Zhang, Pan, Jiang, Zou, Wang, Chen, and Ren]{lin2020learning_event_video_deblur}
Songnan Lin, Jiawei Zhang, Jinshan Pan, Zhe Jiang, Dongqing Zou, Yongtian Wang, Jing Chen, and Jimmy Ren.
\newblock Learning event-driven video deblurring and interpolation.
\newblock In \emph{Proc. ECCV}, pages 695--710. Springer, 2020.

\bibitem[Liu et~al.(2023)Liu, An, Liu, Yuan, Chen, Zhou, Li, Wang, and Tian]{liu2023low}
Lin Liu, Junfeng An, Jianzhuang Liu, Shanxin Yuan, Xiangyu Chen, Wengang Zhou, Houqiang Li, Yan~Feng Wang, and Qi Tian.
\newblock Low-light video enhancement with synthetic event guidance.
\newblock In \emph{Proceedings of the AAAI Conference on Artificial Intelligence}, pages 1692--1700, 2023.

\bibitem[Liu et~al.(2021)Liu, Lin, Cao, Hu, Wei, Zhang, Lin, and Guo]{liu2021swin}
Ze Liu, Yutong Lin, Yue Cao, Han Hu, Yixuan Wei, Zheng Zhang, Stephen Lin, and Baining Guo.
\newblock Swin transformer: Hierarchical vision transformer using shifted windows.
\newblock In \emph{Proceedings of the IEEE/CVF international conference on computer vision}, pages 10012--10022, 2021.

\bibitem[Mead(2023)]{mead2023neuromorphic}
Carver Mead.
\newblock Neuromorphic engineering: In memory of misha mahowald.
\newblock \emph{Neural Computation}, 35\penalty0 (3):\penalty0 343--383, 2023.

\bibitem[Mertens et~al.(2007)Mertens, Kautz, and Van~Reeth]{mertens2007exposure}
Tom Mertens, Jan Kautz, and Frank Van~Reeth.
\newblock Exposure fusion.
\newblock In \emph{15th Pacific Conference on Computer Graphics and Applications (PG'07)}, pages 382--390. IEEE, 2007.

\bibitem[Messikommer et~al.(2022)Messikommer, Georgoulis, Gehrig, Tulyakov, Erbach, Bochicchio, Li, and Scaramuzza]{Messikommer_2022_CVPR}
Nico Messikommer, Stamatios Georgoulis, Daniel Gehrig, Stepan Tulyakov, Julius Erbach, Alfredo Bochicchio, Yuanyou Li, and Davide Scaramuzza.
\newblock Multi-bracket high dynamic range imaging with event cameras.
\newblock In \emph{Proceedings of the IEEE/CVF Conference on Computer Vision and Pattern Recognition (CVPR) Workshops}, pages 547--557, 2022.

\bibitem[Nakai et~al.(2013)Nakai, Hoshi, and Taguchi]{nakai2013color2}
Keita Nakai, Yoshikatsu Hoshi, and Akira Taguchi.
\newblock Color image contrast enhacement method based on differential intensity/saturation gray-levels histograms.
\newblock In \emph{2013 International Symposium on Intelligent Signal Processing and Communication Systems}, pages 445--449. IEEE, 2013.

\bibitem[Pan et~al.(2019)Pan, Scheerlinck, Yu, Hartley, Liu, and Dai]{edi_pan}
Liyuan Pan, Cedric Scheerlinck, Xin Yu, Richard Hartley, Miaomiao Liu, and Yuchao Dai.
\newblock Bringing a blurry frame alive at high frame-rate with an event camera.
\newblock In \emph{Proc. CVPR}, pages 6820--6829, 2019.

\bibitem[Patrick et~al.(2008)Patrick, Posch, and Delbruck]{patrick2008128x}
Lichtsteiner Patrick, Christoph Posch, and Tobi Delbruck.
\newblock A 128{\texttimes}128 120 {dB} 15$\mu$ s latency asynchronous temporal contrast vision sensor.
\newblock \emph{IEEE Journal of Solid-State Circuits}, 2008.

\bibitem[Pizer et~al.(1987)Pizer, Amburn, Austin, Cromartie, Geselowitz, Greer, ter Haar~Romeny, Zimmerman, and Zuiderveld]{pizer1987adaptive}
Stephen~M Pizer, E~Philip Amburn, John~D Austin, Robert Cromartie, Ari Geselowitz, Trey Greer, Bart ter Haar~Romeny, John~B Zimmerman, and Karel Zuiderveld.
\newblock Adaptive histogram equalization and its variations.
\newblock \emph{Computer vision, graphics, and image processing}, 39\penalty0 (3):\penalty0 355--368, 1987.

\bibitem[Shang et~al.(2021)Shang, Ren, Zou, Ren, Luo, and Zuo]{shang2021bringing}
Wei Shang, Dongwei Ren, Dongqing Zou, Jimmy~S Ren, Ping Luo, and Wangmeng Zuo.
\newblock Bringing events into video deblurring with non-consecutively blurry frames.
\newblock In \emph{Proc. ICCV}, pages 4531--4540, 2021.

\bibitem[Sun et~al.(2022)Sun, Sakaridis, Liang, Jiang, Yang, Sun, Ye, Wang, and Gool]{sun2022event}
Lei Sun, Christos Sakaridis, Jingyun Liang, Qi Jiang, Kailun Yang, Peng Sun, Yaozu Ye, Kaiwei Wang, and Luc~Van Gool.
\newblock Event-based fusion for motion deblurring with cross-modal attention.
\newblock In \emph{Proc. ECCV}, pages 412--428. Springer, 2022.

\bibitem[Sun et~al.(2023)Sun, Sakaridis, Liang, Sun, Cao, Zhang, Jiang, Wang, and Van~Gool]{sun2023event}
Lei Sun, Christos Sakaridis, Jingyun Liang, Peng Sun, Jiezhang Cao, Kai Zhang, Qi Jiang, Kaiwei Wang, and Luc Van~Gool.
\newblock Event-based frame interpolation with ad-hoc deblurring.
\newblock In \emph{Proceedings of the IEEE/CVF Conference on Computer Vision and Pattern Recognition (CVPR)}, page 22871. IEEE, 2023.

\bibitem[Tulyakov et~al.(2021)Tulyakov, Gehrig, Georgoulis, Erbach, Gehrig, Li, and Scaramuzza]{tulyakov2021time}
Stepan Tulyakov, Daniel Gehrig, Stamatios Georgoulis, Julius Erbach, Mathias Gehrig, Yuanyou Li, and Davide Scaramuzza.
\newblock Time lens: Event-based video frame interpolation.
\newblock In \emph{Proc. CVPR}, pages 16155--16164, 2021.

\bibitem[Tulyakov et~al.(2022)Tulyakov, Bochicchio, Gehrig, Georgoulis, Li, and Scaramuzza]{tulyakov2022time}
Stepan Tulyakov, Alfredo Bochicchio, Daniel Gehrig, Stamatios Georgoulis, Yuanyou Li, and Davide Scaramuzza.
\newblock Time lens++: Event-based frame interpolation with parametric non-linear flow and multi-scale fusion.
\newblock In \emph{Proc. CVPR}, pages 17755--17764, 2022.

\bibitem[Vaswani et~al.(2017)Vaswani, Shazeer, Parmar, Uszkoreit, Jones, Gomez, Kaiser, and Polosukhin]{vaswani2017attention}
Ashish Vaswani, Noam Shazeer, Niki Parmar, Jakob Uszkoreit, Llion Jones, Aidan~N Gomez, {\L}ukasz Kaiser, and Illia Polosukhin.
\newblock Attention is all you need.
\newblock \emph{NIPS}, 30, 2017.

\bibitem[Venkatanath et~al.(2015)Venkatanath, Praneeth, Bh, Channappayya, and Medasani]{venkatanath2015blind}
Narasimhan Venkatanath, D Praneeth, Maruthi~Chandrasekhar Bh, Sumohana~S Channappayya, and Swarup~S Medasani.
\newblock Blind image quality evaluation using perception based features.
\newblock In \emph{2015 twenty first national conference on communications (NCC)}, pages 1--6. IEEE, 2015.

\bibitem[Vitoria et~al.(2022)Vitoria, Georgoulis, Tulyakov, Bochicchio, Erbach, and Li]{vitoria2022event}
Patricia Vitoria, Stamatios Georgoulis, Stepan Tulyakov, Alfredo Bochicchio, Julius Erbach, and Yuanyou Li.
\newblock Event-based image deblurring with dynamic motion awareness.
\newblock \emph{arXiv preprint arXiv:2208.11398}, 2022.

\bibitem[Wang et~al.(2019)Wang, Zhang, Fu, Shen, Zheng, and Jia]{wang2019underexposed}
Ruixing Wang, Qing Zhang, Chi-Wing Fu, Xiaoyong Shen, Wei-Shi Zheng, and Jiaya Jia.
\newblock Underexposed photo enhancement using deep illumination estimation.
\newblock In \emph{Proceedings of the IEEE/CVF conference on computer vision and pattern recognition}, pages 6849--6857, 2019.

\bibitem[Wang et~al.(2021)Wang, Xu, Fu, Lu, Yu, and Jia]{wang2021seeing}
Ruixing Wang, Xiaogang Xu, Chi-Wing Fu, Jiangbo Lu, Bei Yu, and Jiaya Jia.
\newblock Seeing dynamic scene in the dark: A high-quality video dataset with mechatronic alignment.
\newblock In \emph{Proceedings of the IEEE/CVF international conference on computer vision}, pages 9700--9709, 2021.

\bibitem[Wei et~al.(2018)Wei, Wang, Yang, and Liu]{wei2018deep}
Chen Wei, Wenjing Wang, Wenhan Yang, and Jiaying Liu.
\newblock Deep retinex decomposition for low-light enhancement.
\newblock In \emph{British Machine Vision Conference}, 2018.

\bibitem[Wu et~al.(2022)Wu, Weng, Zhang, Wang, Yang, and Jiang]{wu2022uretinex5}
Wenhui Wu, Jian Weng, Pingping Zhang, Xu Wang, Wenhan Yang, and Jianmin Jiang.
\newblock Uretinex-net: Retinex-based deep unfolding network for low-light image enhancement.
\newblock In \emph{Proceedings of the IEEE/CVF conference on computer vision and pattern recognition}, pages 5901--5910, 2022.

\bibitem[Wu et~al.(2023)Wu, Pan, Wang, Yang, Wei, Li, and Shen]{wu2023learning}
Yuhui Wu, Chen Pan, Guoqing Wang, Yang Yang, Jiwei Wei, Chongyi Li, and Heng~Tao Shen.
\newblock Learning semantic-aware knowledge guidance for low-light image enhancement.
\newblock In \emph{Proceedings of the IEEE/CVF Conference on Computer Vision and Pattern Recognition}, pages 1662--1671, 2023.

\bibitem[Xu et~al.(2022)Xu, Wang, Fu, and Jia]{xu2022snr4}
Xiaogang Xu, Ruixing Wang, Chi-Wing Fu, and Jiaya Jia.
\newblock Snr-aware low-light image enhancement.
\newblock In \emph{Proceedings of the IEEE/CVF conference on computer vision and pattern recognition}, pages 17714--17724, 2022.

\bibitem[Xu et~al.(2023)Xu, Wang, and Lu]{xu2023low7}
Xiaogang Xu, Ruixing Wang, and Jiangbo Lu.
\newblock Low-light image enhancement via structure modeling and guidance.
\newblock In \emph{Proceedings of the IEEE/CVF Conference on Computer Vision and Pattern Recognition}, pages 9893--9903, 2023.

\bibitem[Yang et~al.(2020)Yang, Wang, Fang, Wang, and Liu]{yang2020fidelity}
Wenhan Yang, Shiqi Wang, Yuming Fang, Yue Wang, and Jiaying Liu.
\newblock From fidelity to perceptual quality: A semi-supervised approach for low-light image enhancement.
\newblock In \emph{Proceedings of the IEEE/CVF conference on computer vision and pattern recognition}, pages 3063--3072, 2020.

\bibitem[Yang et~al.(2021)Yang, Wang, Huang, Wang, and Liu]{yang2021sparse}
Wenhan Yang, Wenjing Wang, Haofeng Huang, Shiqi Wang, and Jiaying Liu.
\newblock Sparse gradient regularized deep retinex network for robust low-light image enhancement.
\newblock \emph{IEEE Transactions on Image Processing}, 30:\penalty0 2072--2086, 2021.

\bibitem[Yi et~al.(2023)Yi, Xu, Zhang, Tang, and Ma]{yi2023diff}
Xunpeng Yi, Han Xu, Hao Zhang, Linfeng Tang, and Jiayi Ma.
\newblock Diff-retinex: Rethinking low-light image enhancement with a generative diffusion model.
\newblock In \emph{Proceedings of the IEEE/CVF International Conference on Computer Vision}, pages 12302--12311, 2023.

\bibitem[Yu et~al.(2024)Yu, Ren, Han, Wang, Liang, and Shi]{yu2024eventps}
Bohan Yu, Jieji Ren, Jin Han, Feishi Wang, Jinxiu Liang, and Boxin Shi.
\newblock Eventps: Real-time photometric stereo using an event camera.
\newblock In \emph{Proceedings of the IEEE/CVF Conference on Computer Vision and Pattern Recognition}, pages 9602--9611, 2024.

\bibitem[Yu et~al.(2023)Yu, Wang, Zhang, Zhang, Yang, Liu, and Xia]{yu2023learning}
Lei Yu, Bishan Wang, Xiang Zhang, Haijian Zhang, Wen Yang, Jianzhuang Liu, and Gui-Song Xia.
\newblock Learning to super-resolve blurry images with events.
\newblock \emph{IEEE Transactions on Pattern Analysis and Machine Intelligence}, 2023.

\bibitem[Zamir et~al.(2022)Zamir, Arora, Khan, Hayat, Khan, and Yang]{zamir2022restormer}
Syed~Waqas Zamir, Aditya Arora, Salman Khan, Munawar Hayat, Fahad~Shahbaz Khan, and Ming-Hsuan Yang.
\newblock Restormer: Efficient transformer for high-resolution image restoration.
\newblock In \emph{Proceedings of the IEEE/CVF conference on computer vision and pattern recognition}, pages 5728--5739, 2022.

\bibitem[Zhang and Yu(2022)]{zhang2022unifying}
Xiang Zhang and Lei Yu.
\newblock Unifying motion deblurring and frame interpolation with events.
\newblock In \emph{Proc. CVPR}, pages 17765--17774, 2022.

\bibitem[Zhang et~al.(2019)Zhang, Zhang, and Guo]{zhang2019kindling13}
Yonghua Zhang, Jiawan Zhang, and Xiaojie Guo.
\newblock Kindling the darkness: A practical low-light image enhancer.
\newblock In \emph{Proceedings of the 27th ACM international conference on multimedia}, pages 1632--1640, 2019.

\bibitem[Zhou et~al.(2023)Zhou, Teng, Han, Liang, Xu, Cao, and Shi]{zhou2023deblurring}
Chu Zhou, Minggui Teng, Jin Han, Jinxiu Liang, Chao Xu, Gang Cao, and Boxin Shi.
\newblock Deblurring low-light images with events.
\newblock \emph{International Journal of Computer Vision}, 131\penalty0 (5):\penalty0 1284--1298, 2023.

\end{thebibliography}
}

\end{document}